\documentclass[journal]{IEEEtran}
\usepackage{amssymb}
\usepackage{amsmath}
\usepackage{graphicx}
\usepackage{subfigure}
\usepackage{multicol} 
\usepackage{arydshln}
\usepackage{array}
\usepackage{cite}
\usepackage{color}
\newcommand{\RNum}[1]{\uppercase\expandafter{\romannumeral #1\relax}}

\usepackage{arydshln}
\usepackage{times}
\usepackage{soul}
\usepackage{url}
\usepackage{amsthm}
\usepackage{booktabs}
\usepackage{algorithm}
\usepackage{algorithmic}
\usepackage{multirow}
\usepackage{makecell}
\usepackage{subfigure}
\usepackage{caption}
\usepackage{float}
\usepackage{subfloat}

\begin{document}
%
\title{Information-Theoretic Hashing for\\ Zero-Shot Cross-Modal Retrieval}
%
%
%
\author{Yufeng Shi,
        Shujian Yu$^{\dag}$,~\IEEEmembership{Member,~IEEE},
        Duanquan Xu$^{\dag}$,
        Xinge You,~\IEEEmembership{Senior Member,~IEEE},
\thanks{$^{\dag}$To whom correspondence should be addressed.}
\thanks{Y. Shi, D. Xu and X. You are with the School of Electronic Information and Communications, Huazhong University of Science and Technology, Wuhan 430074, China (E-mails: \{yufengshi17, xudq, youxg\}@hust.edu.cn).}
\thanks{S. Yu is with the Machine Learning Group, UiT-The Arctic University of Norway, Troms{\o} 9037, Norway (E-mail: yusj9011@gmail.com).}
\thanks{Manuscript received April 19, 2005; revised August 26, 2015.}}

\markboth{IEEE Transactions on XXX}%
{Shell \MakeLowercase{\textit{et al.}}: Bare Demo of IEEEtran.cls for IEEE Journals}

\maketitle

\begin{abstract}
Zero-shot cross-modal retrieval~(ZS-CMR) deals with the retrieval problem among heterogenous data from unseen classes. Typically, to guarantee generalization, the pre-defined class embeddings from natural language processing (NLP) models are used to build a common space. In this paper, instead of using an extra NLP model to \emph{define} a common space beforehand, we consider a totally different way to \emph{construct} (or learn) a common hamming space from an information-theoretic perspective. We term our model the Information-Theoretic Hashing~(ITH), which is composed of two cascading modules: an Adaptive Information Aggregation~(AIA) module; and a Semantic Preserving Encoding~(SPE) module. Specifically, our AIA module takes the inspiration from the Principle of Relevant Information (PRI) to construct a common space that adaptively aggregates the intrinsic semantics of different modalities of data and filters out redundant or irrelevant information. On the other hand, our SPE module further generates the hashing codes of different modalities by preserving the similarity of intrinsic semantics with the element-wise Kullback–Leibler~(KL) divergence. A total correlation regularization term is also imposed to reduce the redundancy amongst different dimensions of hash codes. Sufficient experiments on three benchmark datasets demonstrate the superiority of the proposed ITH in ZS-CMR. Source code is available in the supplementary material.
\end{abstract}

\IEEEpeerreviewmaketitle

\section{Introduction}\label{sec1}
The rapid development of consumer electronics caused an exponential growth in the volumes of multi-modal data, which enables humans to apply a sample to seek out relevant data in different modalities. This task is also known as the cross-modal retrieval~\cite{wang2016comprehensive,peng2017overview,9269483}. Recently, cross-modal hashing~\cite{jiang2017deep,ssah,xu2017learning} which adopts hashing functions to map multi-modal data into the common hamming space, has attracted increasing attention from both academia and industry. Compared with the traditional Approximate Nearest Neighbor~(ANN) search algorithms~\cite{wang2015joint,wang2017adversarial,bogolin2022cross}, the hashing methods not only occupy low memory usage due to its compact binary codes, but also enjoy high query speed by making use of Hamming distance.

\begin{figure}[!t]
\centering
\includegraphics[scale=0.6]{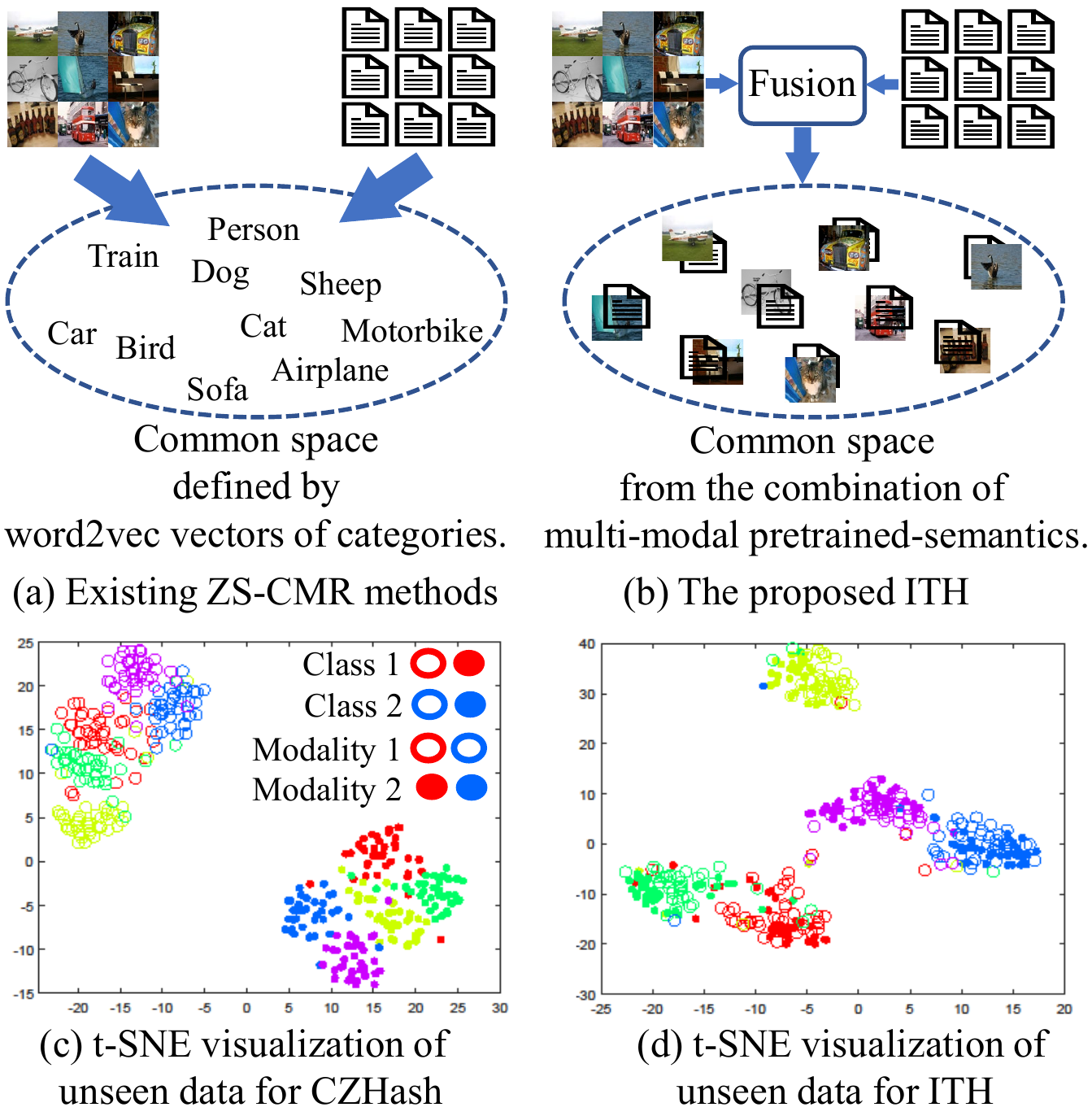}
\caption{Illustration of Information-Theoretic Hashing (ITH). Current methods build the common space based on the pre-defined class embeddings from nature language processing models, such as word2vec~(see (a)). However, However, this common space may not be compatible with the intrinsic semantics of multi-modal data, and the modality gap still exists (see (c)). Conversely, our ITH aims to learn a common space by preserving the multi-modal semantics from an information-theoretic perspective (see (b)). It adaptively fills in the modality gap (see (d)).}\label{fig1}
\end{figure}

Data from different modalities usually follow different distributions or have different modes, which is also referred to as the modality gap~\cite{peng2017overview,ssah,8517125}. To fill in the modality gap, researchers try to build a common space with the common semantics of heterogeneous data. Existing cross-modal retrieval methods can be roughly divided into supervised approach~\cite{jiang2017deep,ssah,shi2019equally} and unsupervised approach~\cite{su2019deep,liu2020joint,zhang2021high}. With the availability of manual annotations, supervised approach directly adopts the annotated semantics to build the common hamming space. As a representative method, Deep Cross-Modal Hashing~(DCMH)~\cite{jiang2017deep} adopts class labels to establish an inter-modal similarity matrix $S$, and encodes the heterogeneous data by making inner products between hashing codes match well with $S$. Nevertheless, data labeling involves expensive means in terms of cost and labor time, which is infeasible in real-world applications. On the other hand, unsupervised approach aims to encode data without labels. To build the common hamming space, existing unsupervised methods focus on the estimation of semantics. For example, Deep Joint-Semantics Reconstructing Hashing~(DJSRH)~\cite{su2019deep} trains hash functions with a two-stage strategy. Since the multi-modal features contain rich semantics~\cite{dong2017semantic,Zhang_2020_AAAI,Zhang_2020_CVPR}, DJSRH integrates the neighborhood relations of different modalities into a joint similarity matrix. However, the assumption that classes in the training data and the test data are consistent severely limits the practical usage of existing unsupervised methods. 


For applications in realistic scenarios, zero-shot cross-modal retrieval~(ZS-CMR) emerges as a new challenge. Specifically, ZS-CMR aims to perform retrieval among heterogeneous data of unseen classes by transferring knowledge learnt from multi-modal data in the seen classes. To guarantee generalization, inspired by popular zero-shot image classification approaches~\cite{8643433,xie2021generalized,chen2021hsva}, the class embeddings from pre-trained natural language processing~(NLP) models~(e.g., word2vec~\cite{mikolov2013efficient}) are utilized as an extra guiding signal to define a common space. Therefore, during training, current ZS-CMR methods~\cite{agnet,czhash,masln,danzcr} optimize models to constrain seen data points aggregating around their corresponding pre-defined class embeddings (see Fig.~\ref{fig1}(a)). Later, in the test phase, the well-trained models are also expected to project data points of unseen classes around their corresponding pre-defined class embeddings. In this sense, the success of knowledge transfer from seen classes to unseen classes is mainly determined by the quality or the transferability of the pre-defined common space.


In this paper, we argue and also empirically demonstrate that the modality gap still exists for state-of-the-art (SOTA) ZS-CMR methods based on pre-defined class embeddings~(see, Fig.~\ref{fig1}(c)).
In fact, these methods implicitly assume that the semantics of any multi-modal data can always be represented by their pre-defined class embeddings, which, from our perspective, could be violated in practice. 
For example, ``frisbee'' is visually similar to ``plate'', but their usages are totally different, which leads to disagreements in texts. Meanwhile, ``plate'' is often accompanied by ``fork'' in articles, but their visual appearance are not similar. Moreover, different languages and NLP models may lead to divergent or inconsistent relations for pre-defined class embeddings. What is worse, new classes may not be included in previously trained NLP models, which makes most of existing ZS-CMR methods become immediately infeasible. For example, the disease ``Cardiomegaly'' cannot be vectorized by the word2vec model in $2013$. For reasons above, the pre-defined common space might be biased to partial data, resulting in modality gap. 




To address the potentially incompatibility between a pre-defined common space and the semantics of heterogeneous data, we investigate the feasibility to construct or learn a common space from given data, without the guidance from an extra NLP model. Our key ideas are straightforward. As the infrastructure in artificial intelligence, pre-trained models for various modalities have been built. Recent examples include the TERA in speech~\cite{liu2021tera} and the MaskFeat for videos~\cite{wei2022masked}. 
With large network structures with billions of parameters and massive training data, features extracted by these pre-trained models contain rich semantics and demonstrate appealing performances in various tasks within the corresponding modality. Therefore, how to precisely inherit or \emph{preserve} the semantics of the features from these individually pre-trained models and how to \emph{reduce} the semantic uncertainty or irrelevant information for specific datasets are the key to boost the generalization of hash codes in ZS-CMR. 

Albeit easy to understand, the information ``preservation" and ``reduction" seem to be less tractable and hard to implement. To this end, we leverage the basic concepts from Shannon's Information Theory~\cite{cover1999elements} and show that information-theoretic measures and principles provide an elegant language to describe both terms with strong theoretical guarantee. For example, the divergence can statistically measures the amount of preserved information from a distribution discrepancy perspective; whereas the entropy directly quantify the extent of information reduction\footnote{A zero entropy means no uncertainty and occurs if and only if all data points are converged to a single point.}. 

Motivated by recent advances in Information-Theoretic Learning~\cite{principe2010information}, we develop Information-Theoretic Hashing (ITH), a novel ZS-CMR model that learns to construct a common hamming space from training data without the guidance from any extra NLP model. The pipeline of our proposed ITH is illustrated in Fig.~\ref{fig2}, which mainly consists of two cascading modules: (1) an Adaptive Information Aggregation~(AIA) model; and (2) a Semantic Preserving Encoding~(SPE) model. Specifically, AIA adaptively aggregates the rich multi-modal semantics into a common continuous space by taking inspiration from the Principle of Relevant Information (PRI)~\cite{principe2010information,li2020pri}, whereas SPE further transform continuous code into binary code by preserving intrinsic semantics. 



To summarize, our main contributions include:
\begin{itemize}
\item~To the best our knowledge, ITH is the first zero-shot cross-modal retrieval (ZS-CMR) model that is designed from an information-theoretic perspective. It adaptively learns a common space from training data to fill in the modality gap (see Fig.~\ref{fig1}(d)) and can be efficiently optimized in an end-to-end manner.


\item~We extend the original PRI from unsupervised formulation to a supervised scenario, such that it can be naturally integrated within the AIA module to aggregate semantics of different modalities, without the guidance of an extra NLP model.

\item~The SPE module elegantly transforms continuous codes into Hash codes by an element-wise Kullback-Leibler (KL) divergence. Moreover, a total correlation regularization term is imposed to further reduce redundancy amongst different dimensions of Hash codes.  


\item~Comprehensive experiments on three public benchmarks demonstrate the superior performance of ITH against other ZS-CMR methods.
\end{itemize}


\section{Related work}\label{sec2}
In this section, we first briefly review representative cross-modal hashing methods and float-value zero-shot cross-modal retrieval methods. Next, we introduce the basic elements of Information Theory and the objective of PRI. 


\subsection{Cross-modal retrieval methods}
Based on the availability of semantics labels, existing cross-modal hashing methods include unsupervised and supervised ones. With annotated labels, various relations among heterogeneous data are explored by supervised methods~\cite{shen2020exploiting,8954946,sun2022deep}. Semantic correlation maximization~(SCM)~\cite{zhang2014large} constructs the pairwise semantic similarity with the semantic label vectors. To increase semantics coverage, Semantics-Preserving Hashing~(SePH)~\cite{lin2015semantics} transforms the label affinities into a probability distribution. Meanwhile, since the class labels indicate the class-level relations, the class information is also preserved. Self-Supervised Adversarial Hashing~(SSAH)~\cite{ssah} further aligns the hash codes and their class labels with linear classifiers.

To get rid of manual annotations, unsupervised methods become increasingly popular~\cite{huang2018mhtn,su2019deep,hu2020creating,yu2021deep}. As a pioneer work, Robust and Flexible Discrete Hashing~(RFDH)~\cite{wang2017robust} adopts linear embedding to learn unified codes for different modalities of one multimodal instance. To extra reveal properties of specific modalities, Joint and individual matrix factorization hashing~(JIMFH)~\cite{wang2020joint} combines unified and individual features to obtain the final hash codes. Nowadays, large-scale pretrained models become reliable semantics sources. Based on the distance relations of features, Joint-modal Distribution-based Similarity Hashing~(JDSH)~\cite{liu2020joint} constructs a joint similarity matrix to fuse cross-modal similarities, which supervises the hashing function learning. High-order Nonlocal Hashing~(HNH)~\cite{zhang2021high} merges the local and nonlocal similarities into a joint similarity matrix. Despite the superiority in handling semantics, the assumption that the classes of testing data could be observed in training stage is still away from the real world.


In case of the classes of training data and test data are disjoint, zero-shot cross-modal hashing emerges an a new challenge. As the first attempt, Attribute-guided Network~(AgNet)~\cite{agnet} first aligns different modal data in the pre-defined class embedding space, and then obtains hash codes based on the similarities among pre-defined embeddings. Meanwhile, Cross-modal Zero-shot Hashing~(CZHash)~\cite{czhash} captures the relations between seen and unseen classes by guiding the deep feature mapping into the pre-defined class embedding space. 

Since hashing-based ZS-CMR is still at the start stage, recent float-value ZS-CMR methods are also introduced. As like aforementioned hashing-based ZS-CMR methods, the pre-defined class embeddings are also utilized as side information to transfer knowledge in the common space. And their main difference is how to align multi-modal data and pre-defined class embeddings. For example,  Modal-adversarial Semantic Learning Network~(MASLN)~\cite{masln} and Dual Adversarial Networks for Zero-shot Cross-media Retrieval~(DANZCR)~\cite{danzcr} align the distributions of different modalities via reconstructing multi-modal data. In this procedure, the pre-defined class embeddings are shared by the representation space of both modalities. Later, based on DANZCR, Dual Adversarial Distribution Network~(DADN)~\cite{dadn} further adopts the maximum mean discrepancy criterion to enhance distribution matching between common embeddings and class embeddings. To improve the quality of alignment, advanced network architectures such as variational autoencoders~(e.g., LCALE~\cite{lcale}) and the combination of AutoEncoder and Generative Adversarial Network~(e.g., AAEGAN~\cite{aaegan}) are also proposed. Meanwhile, Ternary Adversarial Networks with Self-supervision~(TANSS)~\cite{tanss} encodes the pre-defined class embeddings to explicitly supervise modality-specific feature learning process. In addition, the pre-defined class embeddings are also regarded as the seeds of multi-modal data. Therefore, Correlated Feature Synthesis and Alignment~(CFSA)~\cite{cfsa} utilizes pre-defined class embeddings to synthesize multi-modal features with semantic correlation, and further aligns synthetic and true features to obtain the common space. 



\subsection{Background Knowledge on Information Theory}

We briefly introduce the basic elements of information theory, such as the concepts of entropy, divergence, mutual information as well as their physical meanings. we also review the objective of the principle of relevant information (PRI)~\cite{principe2010information,li2020pri} - a less well-known unsupervised information-theoretic learning principles. Interested readers can refer to \cite{yu2021information} for a comprehensive survey on relations between different information-theoretic measures and learning principles, such as the information bottleneck~\cite{tishby2000information,tishby2015deep}. 

\subsubsection{Elements of Information Theory}
Given a random variable $X$ with probability density function~(PDF) $f\left( \boldsymbol{x} \right)$ in a finite set $\mathcal{X}$, the information content from observations in $f\left( \boldsymbol{x} \right)$ is measured by $-\log f\left( \boldsymbol{x} \right)$ and the entropy is given by:
\begin{equation}\label{eq_add0}
H\left( \boldsymbol{X} \right) =-\mathbb{E}\left[ \log f\left( \boldsymbol{x} \right) \right] =-\int_{\mathcal{X}}{f\left( \boldsymbol{x} \right) \log f\left( \boldsymbol{x} \right) dx}.
\end{equation}

This definition is due to Shannon~\cite{shannon1948mathematical}, which is a measure of average information and uncertainty. 


Kullback and Leibler~\cite{kullback1951information} generalized Shannon's entropy definition to measure how one probability distribution $f(\boldsymbol{x})$ is different from a reference distribution $g(\boldsymbol{x})$. They proposed such a measure, subsequently called the Kullback-Leibler (KL) divergence or the relative entropy, which is given by:
\begin{equation}\label{eq_add2}
D_{\text{KL}}\left( f||g \right) =\int{f\left( \boldsymbol{x} \right) \log \frac{f\left( \boldsymbol{x} \right)}{g\left( \boldsymbol{x} \right)}dx}.
\end{equation}

The KL divergence can be used to measure the independence between $p$ random variables $\mathbf{x}_1,\mathbf{x}_2,\cdots,\mathbf{x}_p$: if they are stochastically independent, we have $f(\mathbf{x}_1,\mathbf{x}_2,\cdots,\mathbf{x_p}) = f(\mathbf{x}_1)f(\mathbf{x}_2)\cdots f(\mathbf{x}_p)$. 
The KL divergence between joint distribution $f(\mathbf{x}_1,\mathbf{x}_2,\cdots,\mathbf{x_p})$ and the product of marginal distributions $f(\mathbf{x}_1)f(\mathbf{x}_2)\cdots f(\mathbf{x}_p)$ is called the total correlation (TC)~\cite{watanabe1960information}. That is\footnote{Detailed proofs are shown in the supplementary material.}, 
\begin{equation}
\begin{split}
    & TC(\mathbf{x}_1;\mathbf{x}_2;\cdots;\mathbf{x}_p) \\
    & = D_{\text{KL}}\left( f(\mathbf{x}_1,\mathbf{x}_2,\cdots,\mathbf{x}_p)||f(\mathbf{x}_1)f(\mathbf{x}_2)\cdots f(\mathbf{x}_p) \right) \\
    & = \int{f(\mathbf{x}_1,\mathbf{x}_2,\cdots,\mathbf{x_p}) \log \frac{f(\mathbf{x}_1,\mathbf{x}_2,\cdots,\mathbf{x_p})}{f(\mathbf{x}_1)f(\mathbf{x}_2)\cdots f(\mathbf{x}_p)}d\mathbf{x}_1 \cdots d\mathbf{x}_p } \\
    & = \sum_{i=1}^p H(\mathbf{x}_i) - H(\mathbf{x}_1,\mathbf{x}_2,\cdots,\mathbf{x}_p),
\end{split}
\end{equation}
in which $H(\mathbf{x}_1,\mathbf{x}_2,\cdots,\mathbf{x}_p)$ is also called the joint entropy that measures the uncertainty associated with the joint distribution $f(\mathbf{x}_1,\mathbf{x}_2,\cdots,\mathbf{x}_p)$. 

When $p=2$ (i.e., there are only two random variables), total correlation reduces to the popular mutual information:
\begin{equation}
    I(\mathbf{x}_1;\mathbf{x}_2) = H(\mathbf{x}_1) + H(\mathbf{x}_2) - H(\mathbf{x}_1,\mathbf{x}_2).
\end{equation}


\subsubsection{Principle of Relevant Information (PRI)}



\begin{figure}[htbp]
\centering
\subfigure[]{
\begin{minipage}[t]{0.33\linewidth}
\centering
\includegraphics[width=0.95\textwidth]{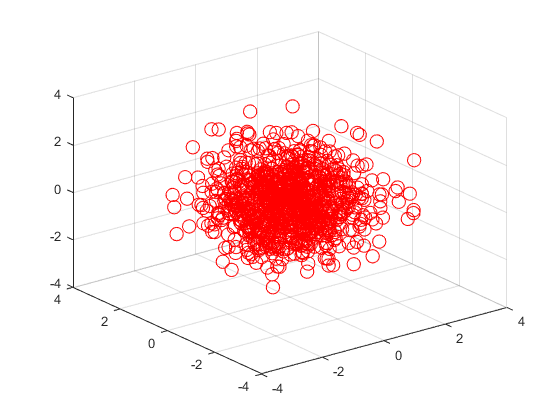}
\end{minipage}%
}%
\subfigure[]{
\begin{minipage}[t]{0.33\linewidth}
\centering
\includegraphics[width=0.95\textwidth]{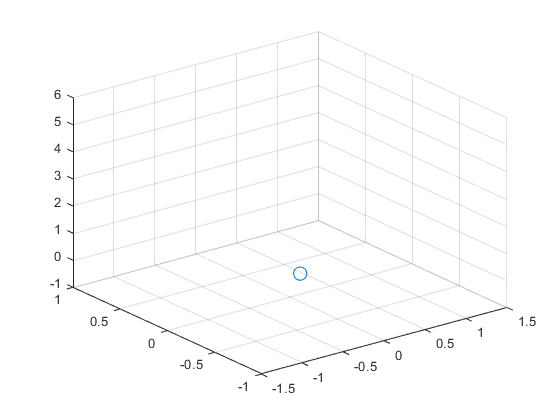}
\end{minipage}%
}%
\subfigure[]{
\begin{minipage}[t]{0.33\linewidth}
\centering
\includegraphics[width=0.95\textwidth]{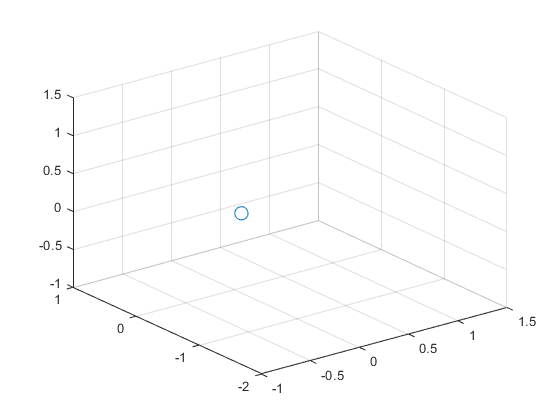}
\end{minipage}%
}%

\subfigure[]{
\begin{minipage}[t]{0.33\linewidth}
\centering
\includegraphics[width=0.95\textwidth]{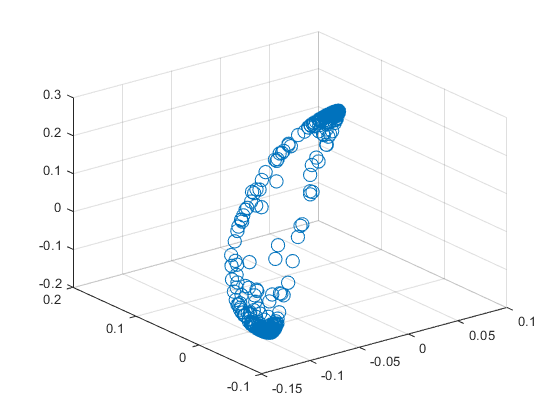}
\end{minipage}%
}%
\subfigure[]{
\begin{minipage}[t]{0.33\linewidth}
\centering
\includegraphics[width=0.95\textwidth]{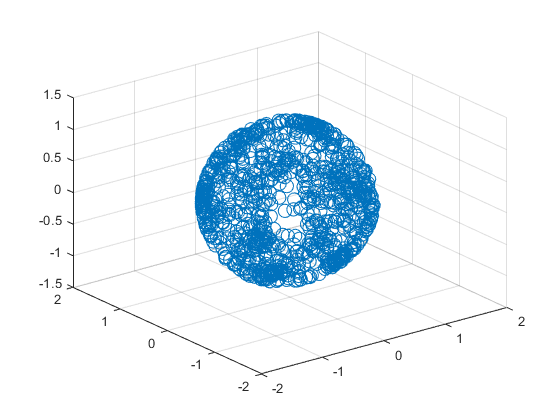}
\end{minipage}%
}%
\subfigure[]{
\begin{minipage}[t]{0.33\linewidth}
\centering
\includegraphics[width=0.95\textwidth]{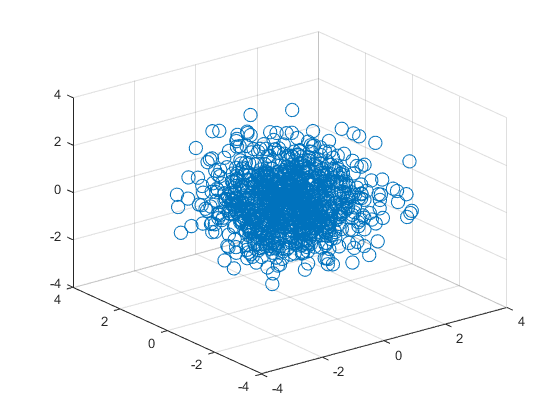}
\end{minipage}%
}%
\centering
\caption{Illustration of the structures revealed by the PRI for (a) a 3d isotropic Gaussian. As the values of $\alpha$ increase, the solution passes through (b) a single point, (c) mode, (d) principal curves, (e) principal surfaces, and in the extreme case of (f) $\alpha\rightarrow\infty$ we get back the data themselves as the solution.}\label{fig_pri_synthetic}
\end{figure}

PRI is a less well-known unsupervised information-theoretic principle that aims to perform mode decomposition of a random variable $\mathbf{x}$ with a known (and fixed) probability distribution $f(\mathbf{x})$. Suppose we
obtain a reduced statistical representation $\mathbf{y}$ with probability distribution $f(\mathbf{y})$. The PRI casts this problem as:
\begin{equation}\label{eq1}
\mathcal{J}_{\text{PRI}}=\min_{\mathbf{y}} H\left( \mathbf{y} \right) +\alpha D\left( f(\mathbf{x})||f(\mathbf{y}) \right),
\end{equation}
where $H\left( \mathbf{y} \right)$ is the entropy of $\mathbf{y}$, $D\left( f(\mathbf{x})||f(\mathbf{y}) \right)$ is the divergence between the distributions of $f(\mathbf{x})$ and $f(\mathbf{y})$, $\alpha$ is a trade-off parameter. The minimization
of entropy can be viewed as a means of reducing uncertainty
and finding the statistical \textbf{regularity} in $\mathbf{y}$, whereas the minimization of divergence ensures the \textbf{descriptive power} of $\mathbf{y}$ about $\mathbf{x}$.
Fig.~\ref{fig_pri_synthetic} illustrates a set of solutions revealed by PRI that are related to the principal curves or surfaces.


In our perspective, the formulation of PRI is compatible with the motivation of aggregation of multi-modal semantics. Specifically, the embeddings of instances should be aligned (with a regular distribution) to reduce uncertainty, and the semantics in different modalities need to be preserved to guarantee descriptive power. 


\begin{figure*}[htp]
\centering
\includegraphics[scale=0.56]{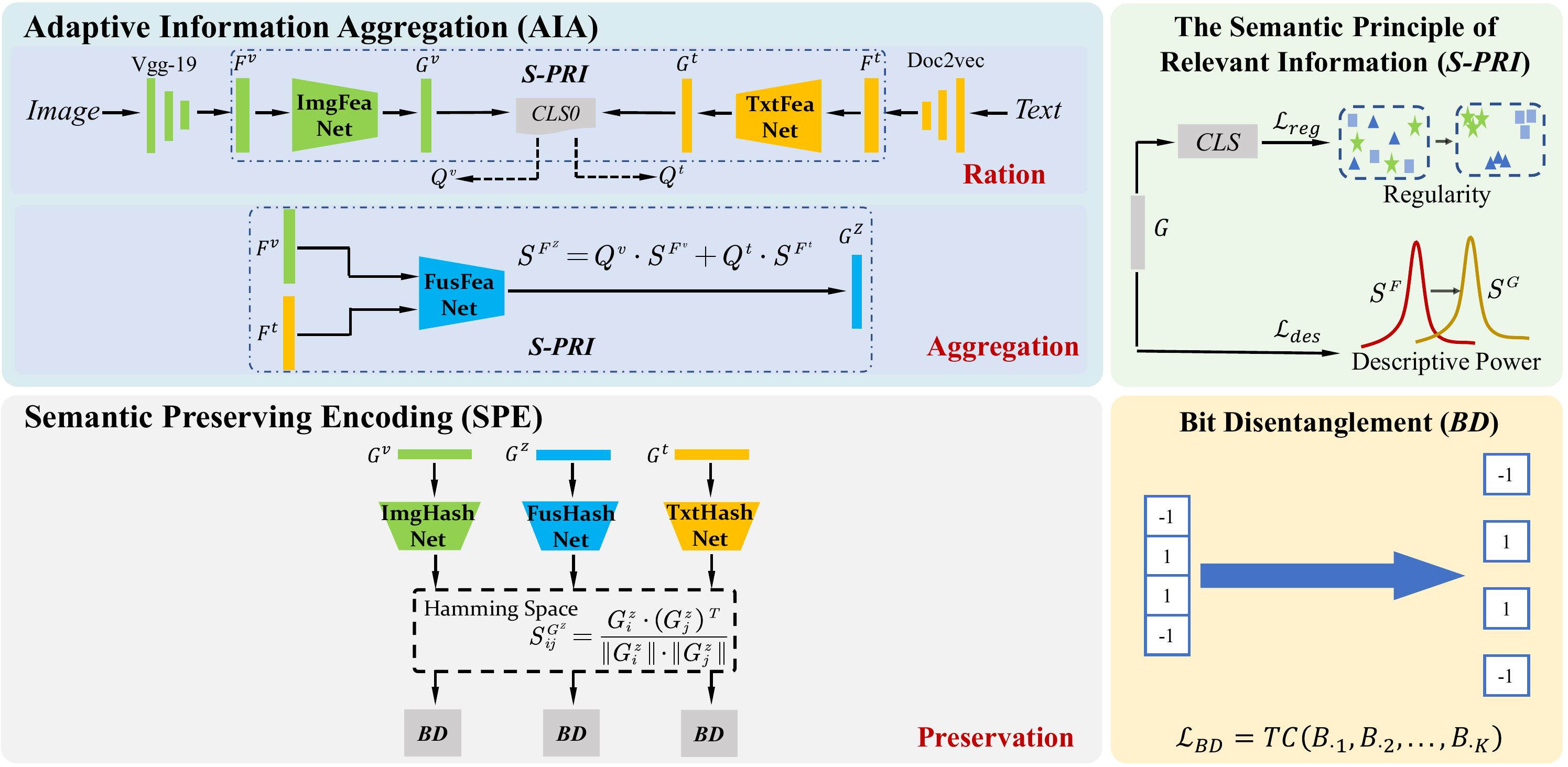}
\caption{A schematic overview of the proposed Information-Theoretic Hashing~(ITH). ITH consists of two cascaded modules, which reduces the modality gap by exploring the intrinsic multi-modal semantics. In AIA, the ration sub-module adopts the common classifier $\textit{CLS0}$ in the S-PRI to quantify the semantics of different modalities, and then the aggregation sub-module integrates the multi-modal semantics with S-PRI. Finally, SPE learns hash codes by preserving the intrinsic multi-modal semantics, and the bit-wise correlations are decoupled to polish hash codes.}\label{fig2}
\end{figure*}


\section{Formulation}\label{sec3}
In this section, Information-Theoretic Hashing~(ITH) is presented in detail. Following typical ZS-CMR methods~\cite{dadn,agnet,cfsa}, ITH performs retrieval between images and texts.

\subsection{Notations and Problem Definition}
Vectors and matrices are denoted by bold lowercase letter~(e.g., $\boldsymbol{x}$) and bold uppercase letter~(e.g., $\boldsymbol{X}$) respectively. $\left\| \boldsymbol{x} \right\| _2$ denotes the 
$2$-norm of vector $\boldsymbol{x}$ and $\left\| \boldsymbol{X} \right\| _F$ denotes the F-norm of matrix $\boldsymbol{X}$.
The sign function is denoted as $sign\left( \cdot \right)$, which outputs $1$ if its input is positive else outputs $-1$. 

In zero-shot scenario, data points used for training and testing are regraded from ``seen classes'' and  ``unseen classes''. They are disjoint~(i.e, $\mathcal{Y}_s\cap \mathcal{Y}_u=\oslash $). Suppose $\mathcal{O}_s=\left\{ o_i \right\} _{i=1}^{N_s}$ denotes $N_s$ multi-modal instances from seen classes, where $o_i=\left( \boldsymbol{x}_{i}^{v},\boldsymbol{x}_{i}^{t},\boldsymbol{y}_i \right) $, $\boldsymbol{x}_{i}^{v}\in \mathbb{R}^v$ and $\boldsymbol{x}_{i}^{t}\in \mathbb{R}^t$ symbolize images and texts\footnote{Because pre-defined class embeddings are not involved in the proposed ITH, corresponding descriptions are omitted.}. Given the code length $K$, cross-modal hashing intends to train hash functions $\mathcal{P}^v\left( \boldsymbol{x}_{i}^{v} \right) :\mathbb{R}^v\rightarrow \left\{ -1,1 \right\} ^K$ and $\mathcal{P}^t\left( \boldsymbol{x}_{i}^{t} \right) :\mathbb{R}^t\rightarrow \left\{ -1,1 \right\} ^K$ with the $N_s$ multi-modal instances from seen classes. In the testing procedure, the well-trained hash functions transform the images and texts of instances from unseen classes~(i.e., $\mathcal{O}_u=\left\{ o_j \right\} _{j=1}^{N_u}$) into hash codes respectively. And the hamming distance $d^B\left( \boldsymbol{B}_{j\cdot}^{v},\boldsymbol{B}_{j\cdot}^{t} \right) $ between hash codes $\boldsymbol{B}_{j\cdot}^{v}$ and $\boldsymbol{B}_{j\cdot}^{t}$ reveals the semantic similarity between $\boldsymbol{x}_{j\cdot}^{v}$ and $\boldsymbol{x}_{j\cdot}^{t}$. To evaluate the retrieval performance, the semantic similarity is defined with annotated labels~\cite{dadn,agnet,cfsa}. For example, if $\boldsymbol{x}_{j\cdot}^{v}$ and $\boldsymbol{x}_{j\cdot}^{t}$ belong to the same class, they should be similar and close to each other in hamming space. Conversely, they are semantically dis-similar. 

\subsection{Information-Theoretic Hashing~(ITH)}
According to aforementioned discussions, proper aggregation and preservation of the intrinsic semantics of multi-modal data is the solution to bridge the modality gap caused by the single-semantic assumption. To implement this idea, Adaptive Information Aggregation~(AIA) and Semantic Preserving Encoding~(SPE) are accordingly developed and integrated in a unified model. Meanwhile, a total correlation regularization  is imposed to further reduce the redundancy among each dimension of hash codes. The unified model is termed as Information-Theoretic Hashing (ITH), whose overview is illustrated in Fig.~\ref{fig2}.

\subsubsection{Adaptive Information Aggregation~(AIA)}
To aggregate the semantics of different modalities, ration and fusion sub-modules are designed. The ration sub-module intends to quantify the semantics, and the aggregation sub-module builds the intrinsic multi-modal semantics.

Before introducing these two sub-modules, we first elaborate a key component inside both sub-modules. We call this component the Semantic Principle of Relevant Information (S-PRI), which is directly motivated by the original formulation of PRI, but introduces label information. 


In zero-shot scenario, the data points of seen classes are given to establish models. Since these training data points are first represented by features of pre-trained models $F$, they mainly own two aspects of semantics: class label and pre-trained semantics. 
To learn a mapping function $\phi$, such that the embeddings $\boldsymbol{G}=\phi(\boldsymbol{F})$ are aligned for different modality of data, we inherit the general idea of PRI and regularize G in such a way that it reduces semantic uncertainty~(in the sense that projected points from the same classes are gathering together) and also has strong descriptive power about $\boldsymbol{F}$. 

For regularity, the semantic uncertainty is reduced by gathering data points from the same classes together:
\begin{equation}\label{eq:regularity}
\mathcal{L}_{reg}=-\log \frac{\exp \left( -\boldsymbol{e}_{i} \right)}{\sum_{j\in \mathcal{Y}_s}{\exp \left( -\boldsymbol{e}_{j} \right)}},
\end{equation}
where $\boldsymbol{e}$ is the classification result of the data representation $G$. For descriptive power, the pre-trained semantics is inherited by minimizing the Kullback-Leibler~(KL) divergence between the semantic distributions of $\boldsymbol{G}$ and $\boldsymbol{F}$. Specifically, the semantic distributions are characterized by the distance relation matrices $\boldsymbol{S}^G$ and $\boldsymbol{S}^F$, with the $(i,j)$-th element defined as:
\begin{equation}\label{eq3}
\boldsymbol{S}_{ij}^{G}=\frac{d^G\left( \boldsymbol{G}_{i\cdot},\boldsymbol{G}_{j\cdot} \right)}{\sum\nolimits_{j\ne i}^{N_s}{d^G\left( \boldsymbol{G}_{i\cdot},\boldsymbol{G}_{j\cdot} \right)}},
\end{equation}
\begin{equation}\label{eq:similarity_F}
\boldsymbol{S}_{ij}^{F}=\frac{d^F\left( \boldsymbol{F}_{i\cdot},\boldsymbol{F}_{j\cdot} \right)}{\sum\nolimits_{j\ne i}^{N_s}{d^F\left( \boldsymbol{F}_{i\cdot},\boldsymbol{F}_{j\cdot} \right)}},
\end{equation}
where $d^G\left( \boldsymbol{G}_{i\cdot},\boldsymbol{G}_{j\cdot} \right)$ indicates the distance between $\boldsymbol{G}_{i\cdot}$ and $\boldsymbol{G}_{j\cdot}$, and $d^F\left( \boldsymbol{F}_{i\cdot},\boldsymbol{F}_{j\cdot} \right)$ indicates the distance between $\boldsymbol{F}_{i\cdot}$ and $\boldsymbol{F}_{j\cdot}$. In our work, the cosine similarity is chosen as the metric for above distances. The term to measure descriptive power can then be expressed with the element-wise KL divergence~\cite{yang2011kullback}:
\begin{equation}\label{eq:element_KL}
\mathcal{L}_{des}=\sum\nolimits_{i=1}^{N_s}{\sum\nolimits_{j=1,j\ne i}^{N_s}{\boldsymbol{S}_{ij}^{G}}}\log \left( \frac{\boldsymbol{S}_{ij}^{G}}{\boldsymbol{S}_{ij}^{F}} \right).
\end{equation}

Combining Eq.~(\ref{eq:regularity}) and Eq.~(\ref{eq:element_KL}), our proposed S-PRI can be formulated as:
\begin{equation}\label{eq6}
\mathcal{L}_{\text{S-PRI}} =\mathcal{L}_{reg}+\alpha \mathcal{L}_{des},
\end{equation}
where $\alpha$ is a trade-off parameter.


To quantify the semantics of different modalities, the ration sub-module utilizes $\textit{ImgFeaNet}(:,\theta _{vg})$ and $\textit{TxtFeaNet}(:,\theta _{tg})$ to transform the original images and texts into the same-dimensional refined representations~(i.e., $\boldsymbol{G}^v$ and $\boldsymbol{G}^t$). Specifically, the $\textit{ImgFeaNet}$ is derived from the VGG-19~\cite{simonyan2014very}, which is pre-trained on the ImageNet dataset. The last layer of VGG-19 is removed to obtain the pre-traind image feature $\boldsymbol{F}^v$, and a one-hidden layer network is added to obtain image refined representation~(i.e., $\boldsymbol{F}^v \rightarrow 4096 \rightarrow \boldsymbol{G}^v$). For $\textit{TxtFeaNet}$, it builds text refined representation $\boldsymbol{G}^t$ from the pre-trained text feature $\boldsymbol{F}^t$ with three fully-connected layers~(i.e., $\boldsymbol{F}^t \rightarrow 4096 \rightarrow \boldsymbol{G}^t$). $\boldsymbol{F}^t$ is extracted by the Doc2vec~\cite{le2014distributed}, which is pre-trained on the English Wikipedia dataset. One should note that the adopted pre-trained models are consistent with current ZS-CMR methods~\cite{danzcr,tanss,aaegan} for fair comparison. 



Following the general idea of S-PRI, we optimize $\theta_{vg}$ and $\theta_{tg}$ separately by the following objective:
\begin{equation}\label{eq7}
\begin{aligned}
\mathcal{L}_{\text{S-PRI}}^{v}=\mathop {arg\min} \limits_{\theta _{vg}}\mathcal{L}_{reg}^{v}+\alpha^{v} \mathcal{L}_{des}^{v},\\
\mathcal{L}_{\text{S-PRI}}^{t}=\mathop {arg\min} \limits_{\theta _{tg}}\mathcal{L}_{reg}^{t}+\alpha^{t} \mathcal{L}_{des}^{t}.
\end{aligned}
\end{equation}

To create the same condition, the semantic uncertainty reduction term in Eq.~(\ref{eq:regularity}) for both image modality and text modality is implemented with the same classifier $\textit{CLS0}(:,\theta _{c0})$.
The hyperparameters $\alpha^v$ and $\alpha^t$ are also remain the same. For $\textit{CLS0}$, it utilizes a two-fully connected layer network (i.e., $G^{\ast} \rightarrow e^{\ast}$, $\ast \in \left\{ v,t \right\} $). To this end, the objective of the ration sub-module is:
\begin{equation}\label{eq8}
\begin{aligned}
\mathcal{L}_{1}=\mathop {arg\min} \limits_{\theta _{vg},\theta _{tg},\theta _{c0}}\mathcal{L}_{\text{S-PRI}}^v + \mathcal{L}_{\text{S-PRI}}^t,
\end{aligned}
\end{equation}

With the same descriptive power, features with strong generalization is easier to be classified, and their corresponding loss value is lower. Therefore, the score of semantics is calculated based on the value of Eq.~(\ref{eq:regularity}):
\begin{equation}\label{eq9}
\begin{aligned}
\boldsymbol{Q}^v=\mathbf{1}-\frac{\mathcal{L}_{reg}^{v}}{\mathcal{L}_{reg}^{v}+\mathcal{L}_{reg}^{t}},\\\boldsymbol{Q}^t=\mathbf{1}-\frac{\mathcal{L}_{reg}^{t}}{\mathcal{L}_{reg}^{v}+\mathcal{L}_{reg}^{t}},
\end{aligned}
\end{equation}
where $\boldsymbol{Q}^v+\boldsymbol{Q}^t=\mathbf{1}$, and $\mathbf{1}$ stands for the $N_s$-dimensional vector with 1 as elements.

After obtaining the score of semantics, the fusion sub-module fuses the semantics of different modalities. Again, following the general idea of S-PRI, the fusion sub-module simply utilizes a one-hidden layer network to estimate the intrinsic semantics from the concatenated features~(i.e., $\textit{FusFeaNet}(:,\theta _{zg}): \left[ \boldsymbol{F}^v;\boldsymbol{F}^t \right] \rightarrow 4096 \rightarrow \boldsymbol{G}^z$). 
With the semantics scores from the well-trained common classifier $\textit{CLS0}$, multi-modal semantics is merged together by re-weighting the original semantic distributions:
\begin{equation}\label{eq10}
\boldsymbol{S}^{F^z}=\boldsymbol{Q}^v\cdot \boldsymbol{S}^{F^v}+\boldsymbol{Q}^t\cdot \boldsymbol{S}^{F^t},
\end{equation}
where $\boldsymbol{S}^{F^v}$ and $\boldsymbol{S}^{F^t}$ are the distance relation matrices calculated by Eq.~(\ref{eq:similarity_F}) using the pretrained features $\boldsymbol{F}^v$ and $\boldsymbol{F}^t$ respectively. Meanwhile, the semantic relations among $G^z$ can be characterized by another matrix $\boldsymbol{S}^{G^z}$ as:
\begin{equation}\label{eq:similarity_fused}
\boldsymbol{S}_{ij}^{G^z}=\frac{d^{G^z}\left( \boldsymbol{G}_{i\cdot}^{z},\boldsymbol{G}_{j\cdot}^{z} \right)}{\sum\nolimits_{j\ne i}^{N_s}{d^{G^z}\left( \boldsymbol{G}_{i\cdot}^{z},\boldsymbol{G}_{j\cdot}^{z} \right)}},
\end{equation}
where $d^{G^z}\left( \boldsymbol{G}_{i\cdot}^{z},\boldsymbol{G}_{j\cdot}^{z} \right)$ indicates the cosine similarity between $\boldsymbol{G}_{i\cdot}^{z}$ and $\boldsymbol{G}_{j\cdot}^{z}$. 

By minimizing the element-wise KL divergence between $S^{F^z}$ and $\boldsymbol{S}^{G^z}$, $\boldsymbol{G}^z$ inherits descriptive power to the concatenated feature $\left[ \boldsymbol{F}^v;\boldsymbol{F}^t \right]$. For regularity, a two-fully connected layer network (i.e., $\textit{CLS1}(:,\theta _{c1}): \boldsymbol{G}^z \rightarrow \boldsymbol{e}^z$) is adopted as the classifier, which reduces the semantic uncertainty. To this end, the objective of the fusion sub-module is:
\begin{equation}\label{eq11}
\begin{aligned}
\mathcal{L}_{2}&=\mathop {arg\min} \limits_{\theta _{zg},\theta _{c1}}\mathcal{L}_{\text{S-PRI}}^{z}\\
&=\mathop {arg\min} \limits_{\theta _{zg},\theta _{c1}}\mathcal{L}_{reg}^{z}+\beta \mathcal{L}_{des}^{z},
\end{aligned}
\end{equation}
where $\beta$ is the hyper-parameter.

Consequently, the intrinsic semantics of multi-modal data is revealed by the fused representation $\boldsymbol{G}^z$.

\subsubsection{Semantic Preserving Encoding~(SPE)}\label{sec_SPE}
By preserving the intrinsic semantics, $\textit{ImgHashNet}(:,\theta _{vb})$ and $\textit{TxtHashNet}(:,\theta _{tb})$ are trained to transform data points as hash codes. Meanwhile, $\textit{FusHashNet}(:,\theta _{zb})$ is also designed, which directly transforms the intrinsic semantics as hash codes to supervise $\textit{ImgHashNet}$ and $\textit{TxtHashNet}$. To better understand the effect of intrinsic semantics, all encoders have the same architecture (i.e., a one-hidden layer network $\boldsymbol{G}^{\ast} \rightarrow 4096 \rightarrow \boldsymbol{B}^{\ast}$, $\ast \in \left\{ v,t,z \right\} $). In the training stage, the requirement of binarization is relaxed to avoid NP-hard problem. As a remedy, $tanh$ that outputs approximate binary codes is adopted as the active function of last layers.

To preserve the intrinsic semantics, the distance relations among hash codes are constrained to reveal the semantic relations among fused representation. The semantic relations among $\boldsymbol{G}^z$ are calculated as Eq.~(\ref{eq:similarity_fused}). Meanwhile, the hamming distance between hash codes can be also calculated by cosine similarity:
\begin{equation}\label{eq13}
d^B\left( \begin{array}{c}	\boldsymbol{B}_{i\cdot},\boldsymbol{B}_{j\cdot}\\\end{array} \right) =\frac{1}{2}\left( c-\boldsymbol{B}_{i\cdot}\left( \boldsymbol{B}_{j\cdot} \right) ^T \right). 
\end{equation}

Therefore, to prevent gradient vanishing in the training stage, the distance relation of hash codes is calculated by cosine similarity:
\begin{equation}\label{eq14}
\boldsymbol{S}_{ij}^{mn}=\frac{c\left( \boldsymbol{B}_{i\cdot}^{m},\boldsymbol{B}_{j\cdot}^{n} \right)}{\sum\nolimits_{j\ne i}^{N_s}{c\left( \boldsymbol{B}_{i\cdot}^{m},\boldsymbol{B}_{j\cdot}^{n} \right)}},
\end{equation}
where $m,n\in \left\{ v,t,z \right\} $ indicate the modalities of hash codes.

Firstly, to achieve effective cross-modal retrieval, inter-modal similarity between different modalities should inherit the intrinsic semantics. Specifically, the inter-modal similarity is represented by the cosine similarity between hash codes from different modalities. By reducing the distribution-wise discrepancy, the inter-modal semantics preservation can be formulated as follows:
\begin{equation}\label{eq15}
\mathcal{L}_{\text{inter}}=\mathop {arg\min} \limits_{\theta _{vb},\theta _{zb},\theta _{tb}}\sum_m{\sum_{n\ne m}{\mathcal{L}_{KL}^{mn}}},
\end{equation}
where $\mathcal{L}_{KL}^{mn}$ indicates the KL divergence between $\boldsymbol{S}^{mn}$ and $\boldsymbol{S}^{G^z}$:
\begin{equation}\label{eq16}
\mathcal{L}_{KL}^{mn}=\mathop {arg\min} \limits_{\theta _{mb},\theta _{nb}}\sum\nolimits_{i=1}^{N_s}{\sum\nolimits_{j=1,j\ne i}^{N_s}{\boldsymbol{S}_{ij}^{mn}}}\log \left( \frac{\boldsymbol{S}_{ij}^{mn}}{\boldsymbol{S}_{ij}^{G^z}} \right).
\end{equation}

Secondly, for image-text pairs that describe the same object, they own common semantics. Therefore, their corresponding hash codes should be aligned. From the pair-wise view, the intra-modal correlation loss is proposed as follows:
\begin{equation}\label{eq17}
\mathcal{L}_{\text{intra}}=\mathop {arg\min} \limits_{\theta _{vb},\theta _{zb},\theta _{tb}}\sum_m{\sum_{n\ne m}{\left\| \boldsymbol{B}^m\left( \boldsymbol{B}^m \right) ^T-\boldsymbol{B}^n\left( \boldsymbol{B}^n \right) ^T \right\| _{F}^{2}}}, 
\end{equation}
where $m,n\in \left\{ v,t,z \right\} $.



Finally, to guarantee the preservation of semantics, the diversity of hash codes should cover the amount of semantics. Therefore, the bit-wise correlation is decoupled, which reduces associated sequences. Specifically, each bit is viewed as a variable, and their total amount of dependence is measured with the Total Correlation~(TC):
\begin{equation}\label{eq18}
TC\left( \boldsymbol{B}_1;\boldsymbol{B}_2;\cdots;\boldsymbol{B}_K \right) =\sum_{i=1}^K H\left( \boldsymbol{B}_{i} \right)  - H(\boldsymbol{B}_1;\boldsymbol{B}_2;\cdots;\boldsymbol{B}_K),
\end{equation}
where $H\left( \cdot \right)$ is the entropy or joint entropy.

In this work, we use the recently proposed matrix-based Renyi’s $\alpha$-order entropy~\cite{yu2019multivariate} to estimate TC, which avoids density estimation in high-dimensional space (see details in the supplementary material). 
The bit-wise uncorrelation loss is defined as:
\begin{equation}\label{eq19}
\mathcal{L}_{\text{tc}}=\mathop {arg\min} \limits_{\theta _{vb},\theta _{zb},\theta _{tb}}TC\left( \boldsymbol{B}^v \right) +TC\left( \boldsymbol{B}^t \right) +TC\left( \boldsymbol{B}^z \right), 
\end{equation}

Combining Eqs.~(\ref{eq15}), (\ref{eq17}) and (\ref{eq19}). the overall objective of SPE is:
\begin{equation}\label{eq20}
\mathcal{L}_3=\mathop {arg\min} \limits_{\theta _{vb},\theta _{zb},\theta _{tb}}\mathcal{L}_{\text{intra}}+\gamma \mathcal{L}_{\text{inter}}+\eta \mathcal{L}_{\text{tc}}, 
\end{equation}
where $\gamma$ and $\eta$ are hyper-parameters.

\subsection{Optimization of ITH}


The ration and fusion sub-modules of AIA and the SPE are trained sequentially. Specifically, for $\textit{ImgFeaNet}$, $\textit{TxtFeaNet}$, $\textit{FusFeaNet}$, $\textit{ImgHashNet}$, $\textit{TxtHashNet}$, $\textit{FusHashNet}$, we use the Adaptive Moment Estimation~(Adam)~\cite{kingma2015adam} optimizer to update their parameters. The detailed optimization procedure of ITH is given in Algorithm~\ref{alg1}.


\begin{algorithm}[tb]
    \caption{The Optimization Procedure of ITH}
    \label{alg1}
    \begin{algorithmic}[1] 
    \renewcommand{\algorithmicrequire}{ \textbf{Input}}
    \REQUIRE{Images $\boldsymbol{X}^v$, texts $\boldsymbol{X}^t$, class labels $\boldsymbol{Y}$, code-length $K$, hyper-parameters $\alpha$,$\beta$, $\gamma$, $\eta$, and batch size $l$.}
    \renewcommand{\algorithmicrequire}{ \textbf{Output}}
    \REQUIRE{Parameters of $\textit{ImgFeaNet}(:,\theta _{vg})$, $\textit{TxtFeaNet}(:,\theta _{tg})$, $\textit{ImgHashNet}(:,\theta _{vb})$ and $\textit{TxtHashNet}(:,\theta _{tb})$.}
    \renewcommand{\algorithmicrequire}{ \textbf{Ration}}
    \REQUIRE{The ration sub-module of AIA}
    \STATE{Initialize $\textit{ImgFeaNet}(:,\theta _{vg})$,  $\textit{TxtFeaNet}(:,\theta _{tg})$, and $\textit{CLS0}(:,\theta _{c0})$.}
    \REPEAT
    \STATE{Randomly sample a batch of $\{( x_{i}^{v},x_{i}^{t},y_i)\}_{i=1}^{l}$ and calculate the corresponding $S^{F^v}$ and $S^{F^t}$ with Eq.~(\ref{eq:similarity_F}).}
    \STATE{Calculate $\mathcal{L}_1$ with Eq.~(\ref{eq8}).}
    \STATE{Update $\theta_{vg}$, $\theta_{tg}$ and $\theta_{c0}$ with $\nabla \mathcal{L}_1$.}
    \UNTIL{convergence.}
    \renewcommand{\algorithmicrequire}{ \textbf{Aggregation}}
    \REQUIRE{The fusion sub-module of AIA}
    \STATE{Initialize $\textit{FusFeaNet}(:,\theta _{zg})$) and $\textit{CLS1}(:,\theta _{c1})$.}
    \REPEAT
    \STATE{Randomly sample a batch of $\{( x_{i}^{v},x_{i}^{t},y_i)\}_{i=1}^{l}$ and calculate the corresponding $S^{F^v}$ and $S^{F^t}$ with Eq.~(\ref{eq:similarity_F}).}
    \STATE{Calculate $Q^v$ and $Q^t$ with Eq.~(\ref{eq9}).}
    \STATE{Calculate $S^{F^z}$ with Eq.~(\ref{eq10}).}
    \STATE{Calculate $\mathcal{L}_2$ with Eq.~(\ref{eq11}).}
    \STATE{Update $\theta _{zg}$ and $\theta _{c1}$ with $\nabla \mathcal{L}_2$.}
    \UNTIL{convergence.}
    \renewcommand{\algorithmicrequire}{ \textbf{Preservation}}
    \REQUIRE{SPE}
    \STATE{Initialize $\textit{ImgHashNet}(:,\theta _{vb})$, $\textit{TxtHashNet}(:,\theta _{tb})$ and $\textit{FusHashNet}(:,\theta _{zb})$.}
    \REPEAT
    \STATE{Randomly sample a batch of $\{(x_{i}^{v},x_{i}^{t}\}_{i=1}^{l}$.}
    \STATE{Extract the refined representations $G^v$, $G^t$ and $G^z$ with the well-trained $\textit{ImgFeaNet}(:,\theta _{vg})$, $\textit{TxtFeaNet}(:,\theta _{tg})$ and $\textit{FusFeaNet}(:,\theta _{zg})$.}
    \STATE{Calculate the common similarity $S^{G^z}$ with Eq.~(\ref{eq:similarity_fused}).}
    \STATE{Calculate $\mathcal{L}_3$ with Eq.~(\ref{eq20}).}
    \STATE{Update $\theta_{vb}$, $\theta_{tb}$ and $\theta_{zb}$ with $\nabla \mathcal{L}_3$.}
    \UNTIL{convergence.}
    \end{algorithmic}
\end{algorithm}

Once Algorithm~\ref{alg1} converges, the well-trained $\textit{ImgHashNet}$ and $\textit{TxtHashNet}$ are adopted to generate hash codes respectively:
\begin{equation}\label{eq21}
\begin{aligned}
\boldsymbol{B}^v=&sign\left( \textit{ImgHashNet}(\boldsymbol{G}^v,\theta _{vb}) \right),\\
\boldsymbol{B}^t=&sign\left( \textit{TxtHashNet}(\boldsymbol{G}^t,\theta _{tb}) \right),
\end{aligned}
\end{equation}
where $\boldsymbol{G}^v$ and $\boldsymbol{G}^t$ symbol the features produced by $\textit{ImgFeaNet}$ and $\textit{TxtFeaNet}$:
\begin{equation}\label{eq21}
\begin{aligned}
\boldsymbol{G}^v=&\textit{ImgFeaNet}(\boldsymbol{F}^v,\theta _{vg}),\\
\boldsymbol{G}^t=&\textit{TxtFeaNet}(\boldsymbol{F}^t,\theta _{tg}).
\end{aligned}
\end{equation}

One should note that for a data point in the test stage, no extra information~(e.g., image-text pairs) is needed to participate in the encoding procedure. Therefore, the proposed ITH is more suitable for realistic applications.

\section{Experiments}\label{sec4}
In this section, extensive experiments on three benchmark datasets are conducted to evaluate and analyze the performance of our proposed ITH. Firstly, Section~\ref{experiment_setting} introduces the adopted datasets and related experiment settings. Then, Section~\ref{performance_eva} demonstrates the performances of ITH and other state-of-the-art (SOTA) competing approaches in terms of hamming ranking and hash lookup. Meanwhile, to thoroughly justify the effectiveness of different modules in ITH, ablation study and parameter analysis are conducted in Section~\ref{ablation_study} and \ref{paramerter_test} respectively. 


\subsection{Experiment Setting}\label{experiment_setting}

\begin{table}[!h]
\caption{Data partitions of three benchmark datasets.}
\centering
\begin{tabular}{|c|c |c c|}
\cline{1-4}
\multirow{2}*{\makecell{\\Dataset}} & Training & \multicolumn{2}{c|}{Testing}\\ 
\cline{2-4}&{\makecell{Seen class\\Database set}}&{\makecell{Unseen class\\Database set}}&{\makecell{Unseen class\\query set}}\\
\cline{1-4}
Wikipedia	&1189	&984	&317\\
Pascal Sentences	&400	&400	&100\\
NUS-WIDE	&24518	&18423	&12158\\
\cline{1-4}
\end{tabular}
\label{tab1}
\end{table}

\begin{table*}[!h]
\caption{Comparison with binary-value baselines in terms of MAP.}
\centering
\scalebox{0.96}[0.96]{
\begin{tabular}{|c | c|c |c c c c |c c c c |c c c c|}
\cline{1-15}
\multirow{2}*{Task} &  \multirow{2}*{Type} & \multirow{2}*{Method} & \multicolumn{4}{c|}{Wikipedia} &  \multicolumn{4}{c|}{Pascal Sentences} &  \multicolumn{4}{c|}{NUS-WIDE}\\ 
\cline{4-15}
        & &  & 16 bits & 32 bits & 64 bits & 128 bits & 16 bits & 32 bits & 64 bits & 128 bits & 16 bits & 32 bits & 64 bits & 128 bits\\
\cline{1-15}
\multirow{11}*{$I\rightarrow T$} & \multirow{5}*{Unsupervised}
& RFDH~\cite{wang2017robust}&0.2670	&0.2550	&0.2634&	0.2595&0.1660	&0.1621	&0.1669	&0.1674&0.4742	&0.4780	&0.4872	&0.5117\\
& & JIMFH~\cite{wang2020joint}&0.2409	&0.2433	&0.2477	&0.2484&0.2029	&0.2298	&0.2658	&0.2717	&0.4604	&0.4681	&0.4972	&0.5100\\
& & DJSRH~\cite{su2019deep}&0.2934	&0.3135	&0.3030	&0.3029&0.3192	&0.3709	&0.4103	&0.4241			&0.4884	&0.5042	&0.5519	&0.5445\\
& & JDSH~\cite{liu2020joint}&0.3109	&0.3456	&0.3557	&0.3506	&0.3219	&0.4154	&0.4440	&0.4661&0.4771	&0.4952	&0.5043	&0.5156\\
& & HNH~\cite{zhang2021high}&0.2555	&0.3115	&0.3149	&0.3181		&0.1875	&0.2383	&0.2388	&0.2384		&0.4747	&0.4861	&0.4737	&0.5122\\
\cdashline{2-15}
& \multirow{4}*{Supervised} & SCM~\cite{zhang2014large} &0.2304	&0.2305	&0.2543	&0.2229	&0.1209	&0.1229	&0.1207	&0.1168	&0.4255	&0.4267	&0.4257	& 0.4045\\
& & SePH~\cite{lin2015semantics} &0.2733	&0.2883	&0.2826	&0.2927	&0.1452	&0.1466	&0.1447	&0.1586	&0.4015	&0.3990	&0.4051	&0.4129\\
& & DCMH~\cite{jiang2017deep} & 0.2267	&0.2205	&0.2243	&0.2192	&0.1189	&0.1232	&0.1202	&0.1152&0.3855	&0.4195	&0.3907	&0.4121\\
& & SSAH~\cite{ssah}&0.2190	&0.2180	&0.2284	&0.2449		&0.1834	&0.1881	&0.1692	&0.1462		&0.4235	&0.3956	&0.4036	&0.3887\\
\cdashline{2-15}
& \multirow{3}*{Zero-shot} & AgNet~\cite{agnet}& 0.2877	&0.2875	&0.2836	&0.2759		&0.3118	&0.3145	&0.3475	&0.3496		&0.4177	&0.4161	&0.4378	&0.4162\\
& & CZHash~\cite{czhash}& 0.2623	&0.2617	&0.2669	&0.2803		&0.2459	&0.2162	&0.2620	&0.2513		&0.3998	&0.4208	&0.4158	&0.4010\\
& & ITH & \textbf{0.3388} &	\textbf{0.3544}&	\textbf{0.3589} &	\textbf{0.3656} &\textbf{0.4201}&	\textbf{0.4583}	&\textbf{0.4645}&	\textbf{0.4755}& \textbf{0.5097}&	\textbf{0.5281}&	\textbf{0.5715}&	\textbf{0.5806}\\
\cline{1-15}
\multirow{11}*{$T\rightarrow I$} & \multirow{5}*{Unsupervised}
& RFDH~\cite{wang2017robust}&0.2610	&0.2548	&0.2474	&0.2500	&0.1635	&0.1679	&0.1780	&0.1548	&0.4714	&0.4857	&0.5062	&0.5256\\
& & JIMFH~\cite{wang2020joint}&0.2464	&0.2428	&0.2451	&0.2481		&0.1828	&0.2108	&0.2334	&0.2466	&0.4353	&0.4649	&0.4744	&0.4747\\
& & DJSRH~\cite{su2019deep}&0.2865	&0.2951	&0.2937	&0.3028	&0.3112	&0.3441	&0.3921	&0.4119		&0.4913	&0.5349	&0.5604	&0.5555\\
& & JDSH~\cite{liu2020joint}&0.3106	&0.3249	&0.3281	&0.3361	&0.3337	&0.3990	&0.4192	&0.4240&0.5134	&0.5458	&0.5463	&0.5532\\
& & HNH~\cite{zhang2021high}&0.2447	&0.2914	&0.3007	&0.2985		&0.1835	&0.1909	&0.1745	&0.2379		&0.4455	&0.4609	&0.4626	&0.5273\\
\cdashline{2-15}
& \multirow{4}*{Supervised} & SCM~\cite{zhang2014large} &0.2182	&0.2206&0.2331&0.2138&0.1229	&0.1142	&0.1150	&0.1120	&0.4298	&0.4332	&0.4336	& 0.3941\\
& & SePH~\cite{lin2015semantics} &0.2490	&0.2561	&0.2587	&0.2600	&0.1881	&0.1501	&0.1467	&0.1826	&0.4297	&0.4316	&0.4392	&0.4507	\\
& & DCMH~\cite{jiang2017deep} & 0.2220	&0.2128	&0.2186	&0.2184	&0.1166	&0.1143	&0.1173	&0.1257&0.3848	&0.4152	&0.3935	&0.4014\\
& & SSAH~\cite{yu2021deep}&0.2198	&0.2180	&0.2161	&0.2272		&0.1892	&0.1892	&0.1892	&0.1216		&0.3984	&0.3746	&0.4082	&0.4072\\
\cdashline{2-15}
& \multirow{3}*{Zero-shot} & AgNet~\cite{agnet}& 0.2482	&0.2481	&0.2524	&0.2463		&0.3123	&0.3001	&0.3160	&0.3066		&0.4105	&0.4039	&0.4454	&0.4421\\
& & CZHash~\cite{czhash}& 0.2566	&0.2587	&0.2709	&0.2694		&0.2297	&0.2321	&0.2700	&0.2601		&0.4432	&0.4646	&0.4713	&0.4627\\
& & ITH & \textbf{0.3168}	&\textbf{0.3333}&	\textbf{0.3403}&	\textbf{0.3403} &\textbf{0.4062}&	\textbf{0.4239}&	\textbf{0.4594}&	\textbf{0.4630} &\textbf{0.5195}&	\textbf{0.5521}&	\textbf{0.5700}&	\textbf{0.5750}\\
\cline{1-15}
\end{tabular}
}
\label{tab2}
\end{table*}

\subsubsection{Dataset Description}
The involved datasets are introduced below, and their corresponding partitions are summarized in Table~\ref{tab1}. Following~\cite{dadn,lcale,aaegan}, 50\% classes are defined as seen classes, and the rest are regarded as unseen classes. The database set of the seen classes are used to train models.  

\textbf{Wikipedia}~\cite{rasiwasia2010new} contains 2866 images that are downloaded from the Wikipedia website. Each image is associated with a text description and a 10-dimensional class label. The text description usually contains several paragraphs. And the class labels indicate the abstract conceptions, such as art and royalty.

\textbf{Pascal Sentences}~\cite{farhadi2010every} is derived from the PASCAL Visual Object Classes Challenge. It consists of 1000 images, which belongs to 20 objects, such as bicycle and cat. For texts, each image are assigned with a document of five sentences.

\textbf{NUS-WIDE}~\cite{chua2009nus} is a classical cross-modal retrieval dataset, which contains 269648 image-tag pairs and corresponding 81-dimensional semantic labels. Since some pairs simultaneously belong to multiple classes, the original dataset is filtered to create the zero-shot scenario. Consequently, 71602 pairs that exclusively belong to the top-10 classes are left for evaluation.

\begin{table*}[!h]
\caption{Comparison with continue-value baselines in terms of MAP.}
\centering
\begin{tabular}{|c |c  c c |c c  c |c c  c|}
\cline{1-10}
\multirow{2}*{Method} & \multicolumn{3}{c|}{Wikipedia} &  \multicolumn{3}{c|}{Pascal Sentences} &  \multicolumn{3}{c|}{NUS-WIDE}\\ 
\cline{2-10}& $I\rightarrow T$ & $T\rightarrow I$ & Avg & $I\rightarrow T$ & $T\rightarrow I$ & Avg & $I\rightarrow T$ & $T\rightarrow I$ & Avg\\
\cline{1-10}
MASLN~\cite{masln} &0.284&	0.264&	0.274&	0.307&	0.294	&0.301&	0.411&	0.426&	0.419\\
DANZCR~\cite{danzcr} &0.297	&0.287	&0.292	&0.334	&0.338	&0.336&	0.416&	0.469&	0.443\\
DADN~\cite{dadn} &0.305	&0.291&	0.298&	0.359&	0.353&	0.356&	0.423&	0.472&	0.448\\
TANSS~\cite{tanss} &0.313&	0.289&	0.301&	0.351&	0.365&	0.358&	0.487&	0.493&	0.490\\
CFSA~\cite{cfsa} &0.341&	0.311&	0.326&	0.378&	0.368&	0.373&	0.501&0.507	&0.504\\
LCALE~\cite{lcale} &0.367&	0.357&	0.362&	0.414&	0.394&	0.404&	0.566&	0.567&	0.567\\
AAEGAN~\cite{aaegan} &0.395	&0.346&	0.371&	0.437&	0.412&	0.425&	0.584	&0.587&	0.586\\
ITH & \textbf{0.367}	&\textbf{0.343}	&\textbf{0.355}	&\textbf{0.494}	&\textbf{0.476}	&\textbf{0.485}	&\textbf{0.597}	&\textbf{0.593}	&\textbf{0.595}\\ 
\cline{1-10}
\end{tabular}
\label{tab3}
\end{table*}

\subsubsection{Evaluation Metric}
To measure the accuracy of hashing-based retrieval, hamming ranking and hash lookup are used as protocols. Hamming ranking is to sort the database set based on the hamming distance between the query point and the retrieved points. Therefore, the Mean Average Precision~(MAP) is used. The MAP is the most common metric in retrieval tasks, which indicates the average precision of returned points. Meanwhile, hash lookup is to return data points within a certain hamming distance radius to the query point. Therefore, the Precision-Recall~(PR) curve is adopted. 

\subsubsection{Implementation Detail}
The proposed ITH is compared with $11$ SOTA hashing-based methods, where RFDH~\cite{wang2017robust} and JIMFH~\cite{wang2020joint} are unsupervised shallow methods, DJSRH~\cite{su2019deep}, JDSH~\cite{liu2020joint} and HNH~\cite{zhang2021high} are unsupervised deep methods, SCM~\cite{zhang2014large} and SePH~\cite{lin2015semantics} are supervised shallow methods, DCMH~\cite{jiang2017deep} and SSAH~\cite{ssah} are supervised deep methods, and AgNet~\cite{agnet} and CZHash~\cite{czhash} are zero-shot deep methods. Sources codes of all competitors are kindly provided. We also inherit the same hyper-parameters as suggested by their corresponding authors. Meanwhile, to comprehensively analyze ITH, seven recently-proposed float-value methods for zero-shot scenarios including DEMZSL~\cite{demzsl}, MASLN~\cite{masln}, DANZCR~\cite{danzcr}, DADN~\cite{dadn}, TANSS~\cite{tanss}, LCALE~\cite{lcale} and AAEGAN~\cite{aaegan} are also selected for comparison in terms of MAP. For the proposed ITH, it is implemented with the Pytorch framework on a server with one 1080-Ti GPU. The model is optimized by the Adam optimizer with learning rate $0.001$. On all datasets, the hyper-parameters of ITH are empirically set as: $\alpha=100,\beta=100,\gamma=1, \eta=0.01$.

To have a fair comparison, following current baselines~\cite{agnet,lcale,aaegan}, the VGG-19~\cite{simonyan2014very} pretrained on ImageNet is used to process images and extract the 4096-dimensional features, and 300-dimensional features of texts are extracted by the Doc2vec~\cite{le2014distributed} pretrained on Wikipedia. Since existing works need extra pre-defined class-embeddings to boost generalization, the Word2vec~\cite{mikolov2013efficient} model, which is pre-trained on Google News, is adopted to build 300-dimensional vectors for class names. Meanwhile, the generalization of ITH is only based on the pretrained image and text features. 

\subsection{Performance Evaluation}\label{performance_eva}
In this part, two frequently-used cross-modal retrieval tasks:1) $I\rightarrow T$: using images to retrieve texts, and 2) $T\rightarrow I$: using texts to retrieve images, are conducted on three benchmark datasets. The retrieval performance of the proposed ITH and all competitors in terms of hamming ranking and hash lookup is reported and analyzed.

\subsubsection{Hamming ranking}
Table~\ref{tab2} reports the MAP of ITH and binary-value competitors on Wikipedia, Pascal Sentences and NUS-WIDE datasets with $16$, $32$, $64$ and $128$ bits of hash codes. From Table~\ref{tab2}, several phenomenons can be observed. Firstly, comparing DCMH and SSAH with AgNet and CZHash, the supervised methods are not valid in the zero shot scenario. With extra considering the knowledge transfer, the zero-shot methods outperform the supervised methods. Secondly, comparing RFDH, JIMFH, DJSRH, JDSH and HNH with other baselines, the unsupervised methods outperform the supervised methods and the zero-shot methods on the whole. Since the unsupervised methods only utilize the pre-trained features to optimize models, the generalization of the pre-trained semantics is reliable. Thirdly, comparing RFDH and JIMFH with DJSRH, JDSH and HNH, deep methods surpass shallow methods, which demonstrates that learning from scratch can favor the retrieval accuracy. 

For the proposed method, ITH outperforms all binary-value competitors in all cases under the zero-shot setting. It not only integrates above-motioned favorable factors~(i.e., using the semantics of pretrained features to guide the hash codes learning from scratch), but also equips with extra novel designs. Compared with the deep unsupervised methods, ITH extra utilizes the class labels under the S-PRI framework. Since the semantic uncertainty is reduced for specific datasets, the hash codes of ITH are more concentrated. Meanwhile, rather than the coarse weights of different modalities from humans, the AIA in ITH adaptively merges the multi-modal semantics with the consideration of generalization. Therefore, the subsequent hash code learning is accompanied by more efficient guidance. Furthermore, current methods also ignore the impact of bit-wise correlation, which weakens the representation ability of hash codes and also introduces potential redundancy (amongst each dimension of hash codes). The performance of AgNet and CZHash on the NUS-WIDE dataset is taken as example, counter-intuitively, the MAPs of $128$ bits are lower than the MAPs of $64$ bits. As for the proposed ITH, it optimizes the Total Correlation of hash codes to reduce the bit-wise correlation, which further boosts the retrieval performance.

To further evaluate the efficiency of ITH, seven recently-proposed float-value methods for ZS-CMR are utilized for extra comparison. Table~\ref{tab3} reports the MAP of the proposed ITH with 256 bits~(following the 256-dim float vector in AAEGAN~\cite{aaegan}) and float-value baselines with their own dimensions on Wikipedia, Pascal Sentences and NUS-WIDE datasets. The corresponding MAPs of the sate-of-the-art methods are reported by~\cite{lcale,aaegan}. By exploring the intrinsic semantics of multi-modal data, the proposed ITH achieves the best accuracy on the Pascal Sentences and NUS-WIDE datasets. In Wikipedia dataset, our ITH ranks the third. Comparing the Wikipedia dataset with the other two datasets, the Wikipedia dataset describes the abstract concepts such as art and royalty, while the other two datasets consist of objects in real world like bike and motorbike. Since the natural language is privileged in describing abstract concepts, together with the complex architectures, it makes sense that the accuracy of LCALE and AAEGAN is higher than ITH in Wikipedia. Meanwhile, with no extra information needed, the simple-structure ITH is more suitable to the practical objects. Consequently, the MAP results confirm the superiority of ITH in terms of the hamming ranking protocol.

\subsubsection{Hash lookup}
\begin{figure}[htbp]
\centering
\subfigure[]{
\begin{minipage}[t]{0.49\linewidth}
\centering
\includegraphics[width=1.7in]{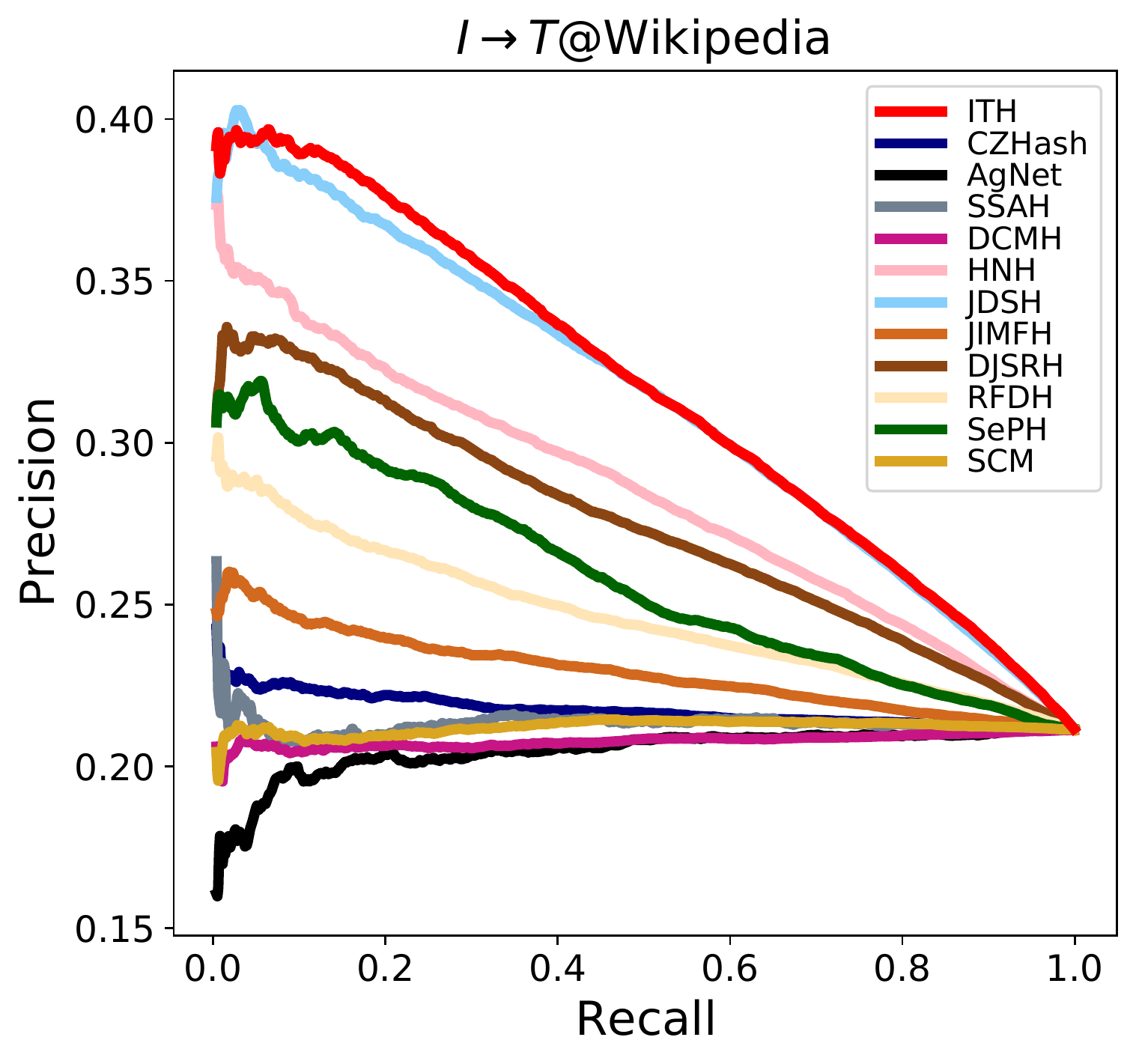}
\end{minipage}%
}%
\subfigure[]{
\begin{minipage}[t]{0.49\linewidth}
\centering
\includegraphics[width=1.7in]{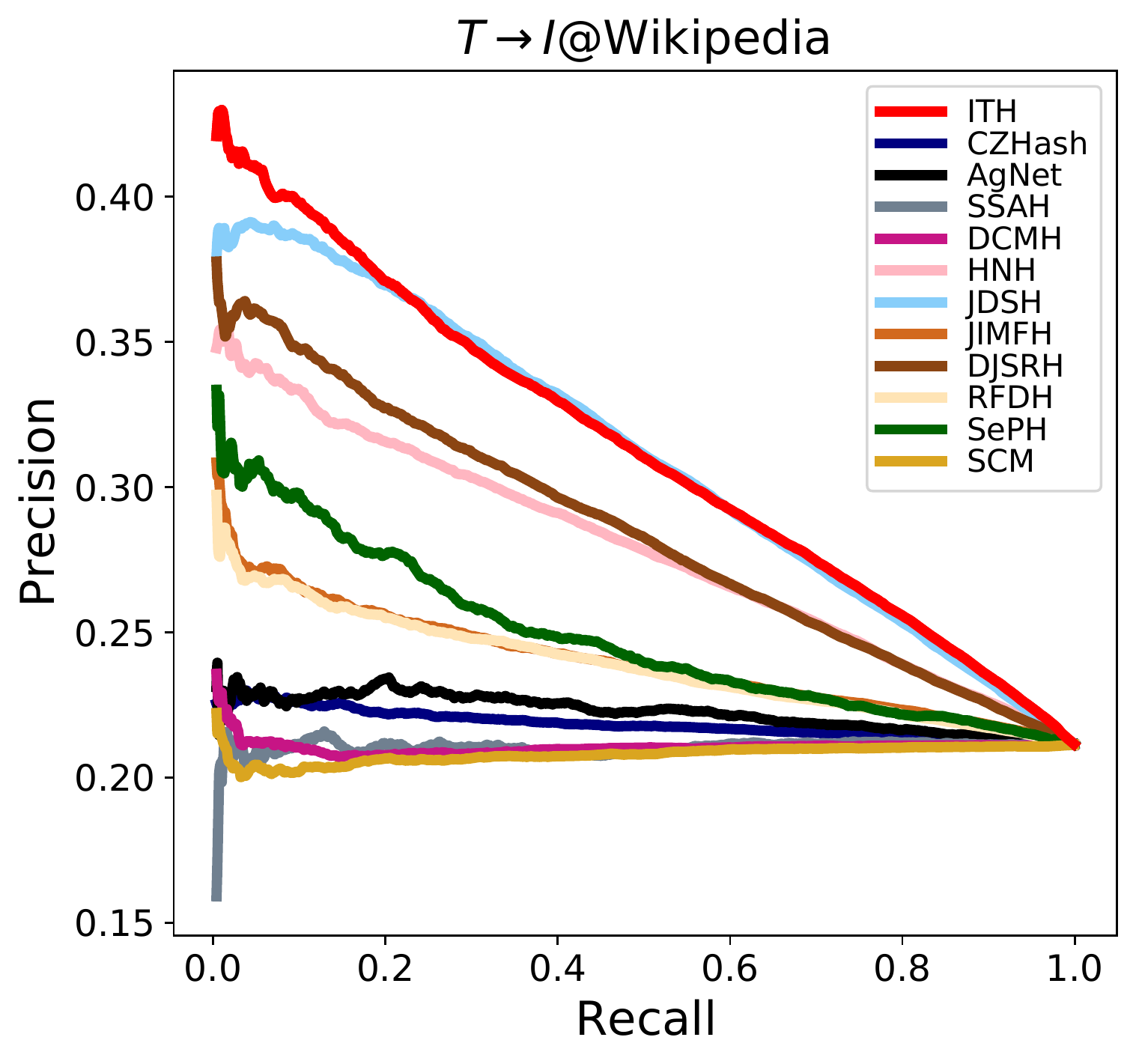}
\end{minipage}%
}%

\subfigure[]{
\begin{minipage}[t]{0.49\linewidth}
\centering
\includegraphics[width=1.7in]{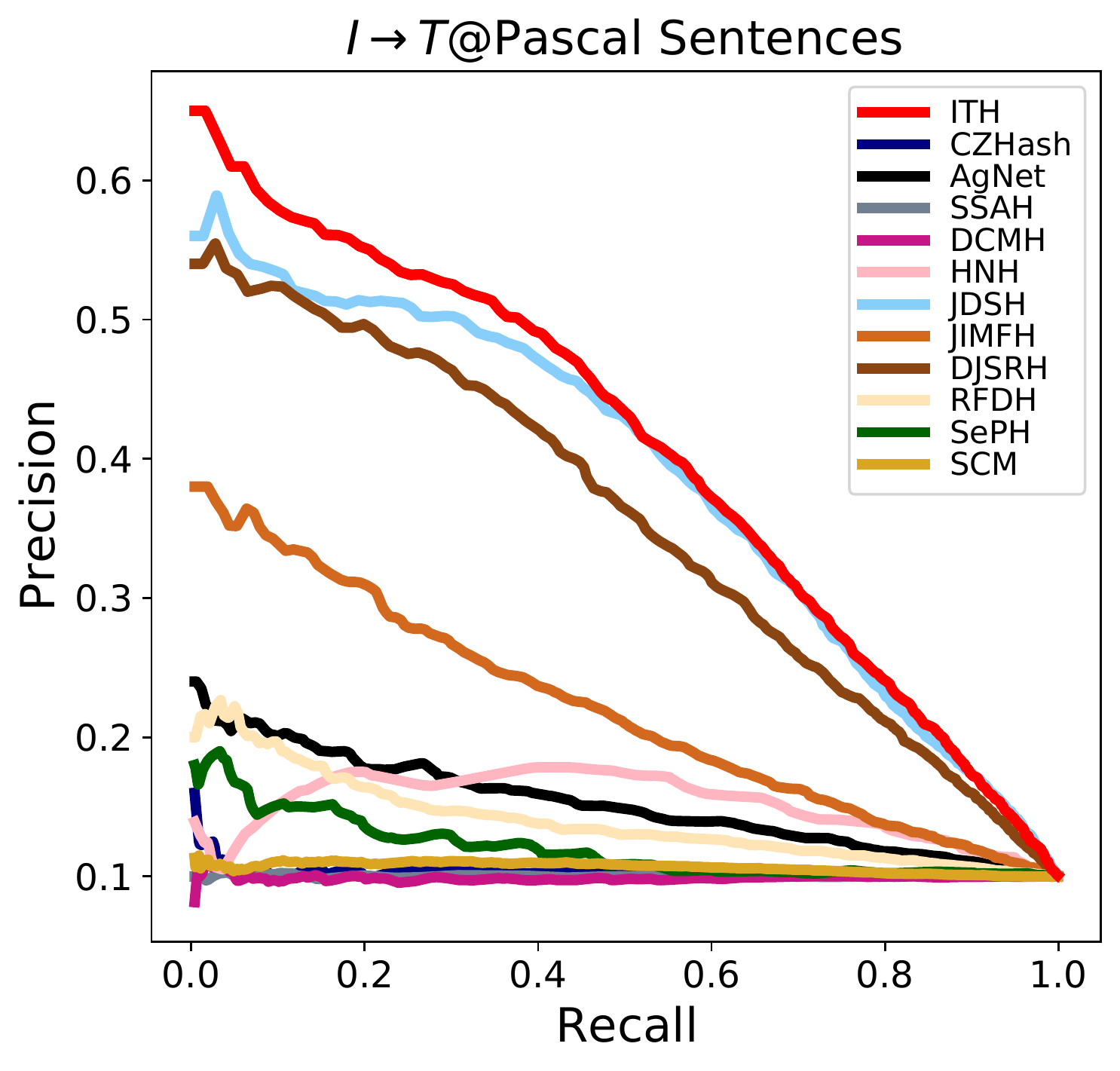}
\end{minipage}%
}%
\subfigure[]{
\begin{minipage}[t]{0.49\linewidth}
\centering
\includegraphics[width=1.7in]{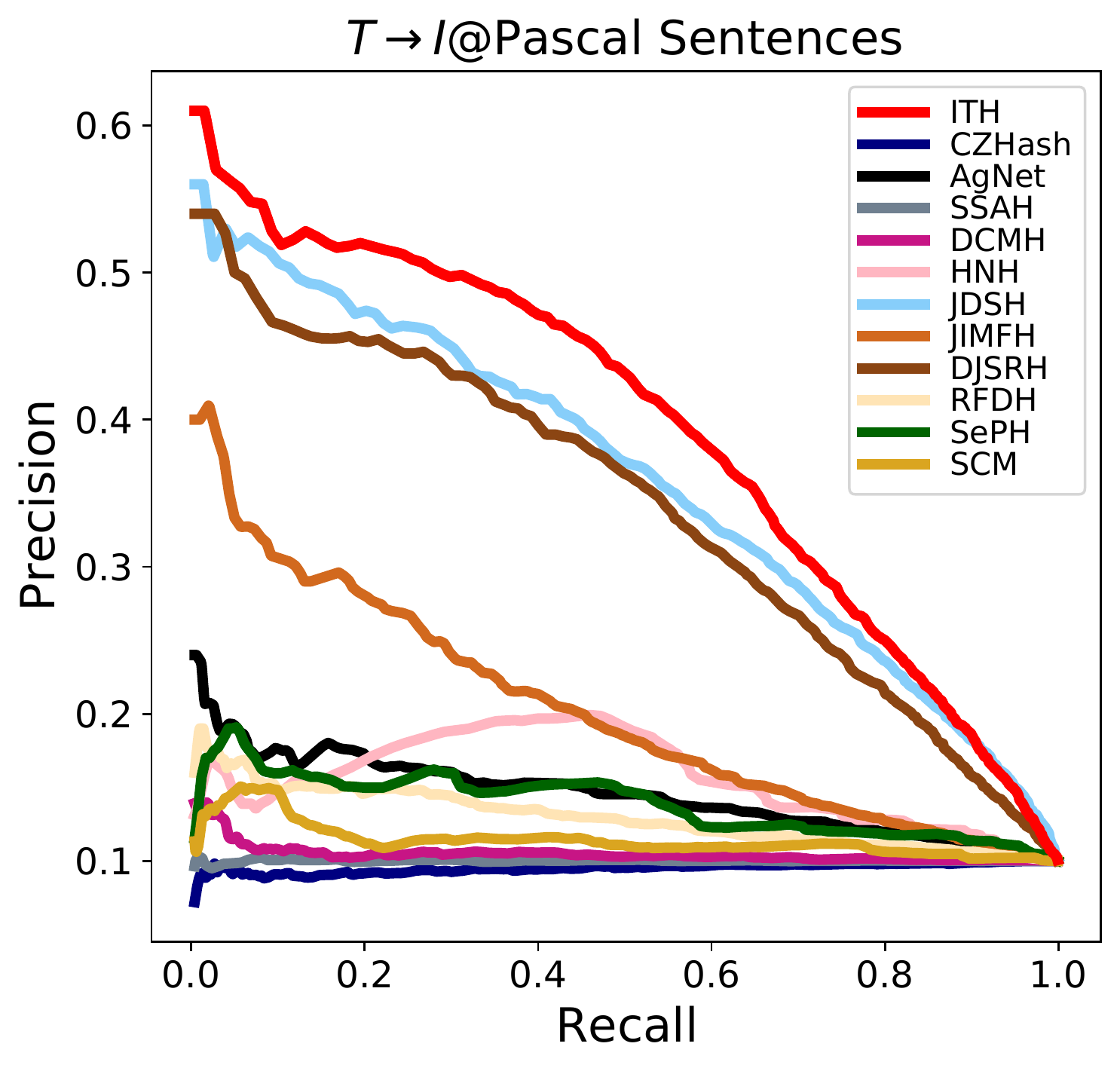}
\end{minipage}%
}%

\subfigure[]{
\begin{minipage}[t]{0.49\linewidth}
\centering
\includegraphics[width=1.7in]{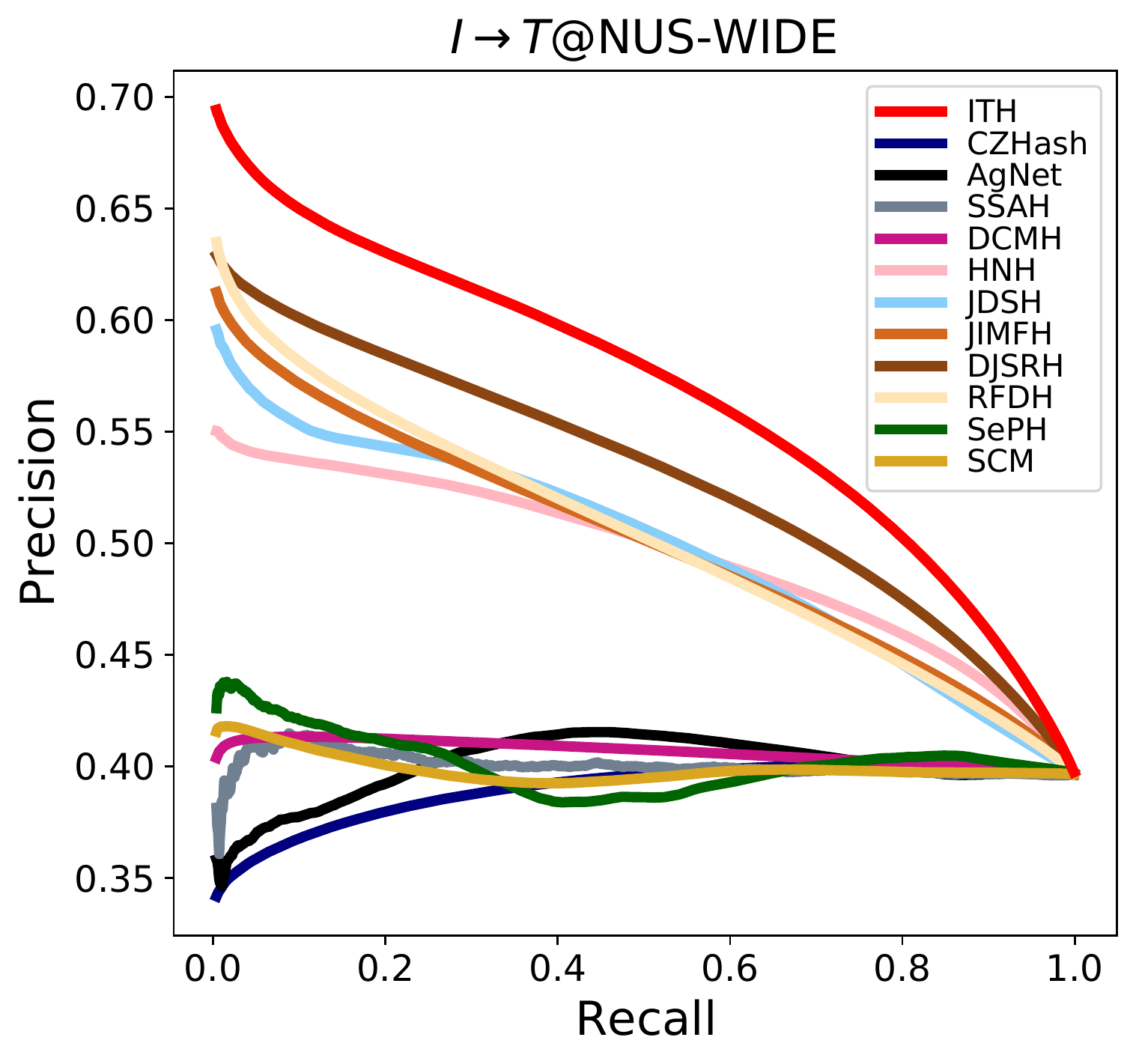}
\end{minipage}%
}%
\subfigure[]{
\begin{minipage}[t]{0.49\linewidth}
\centering
\includegraphics[width=1.7in]{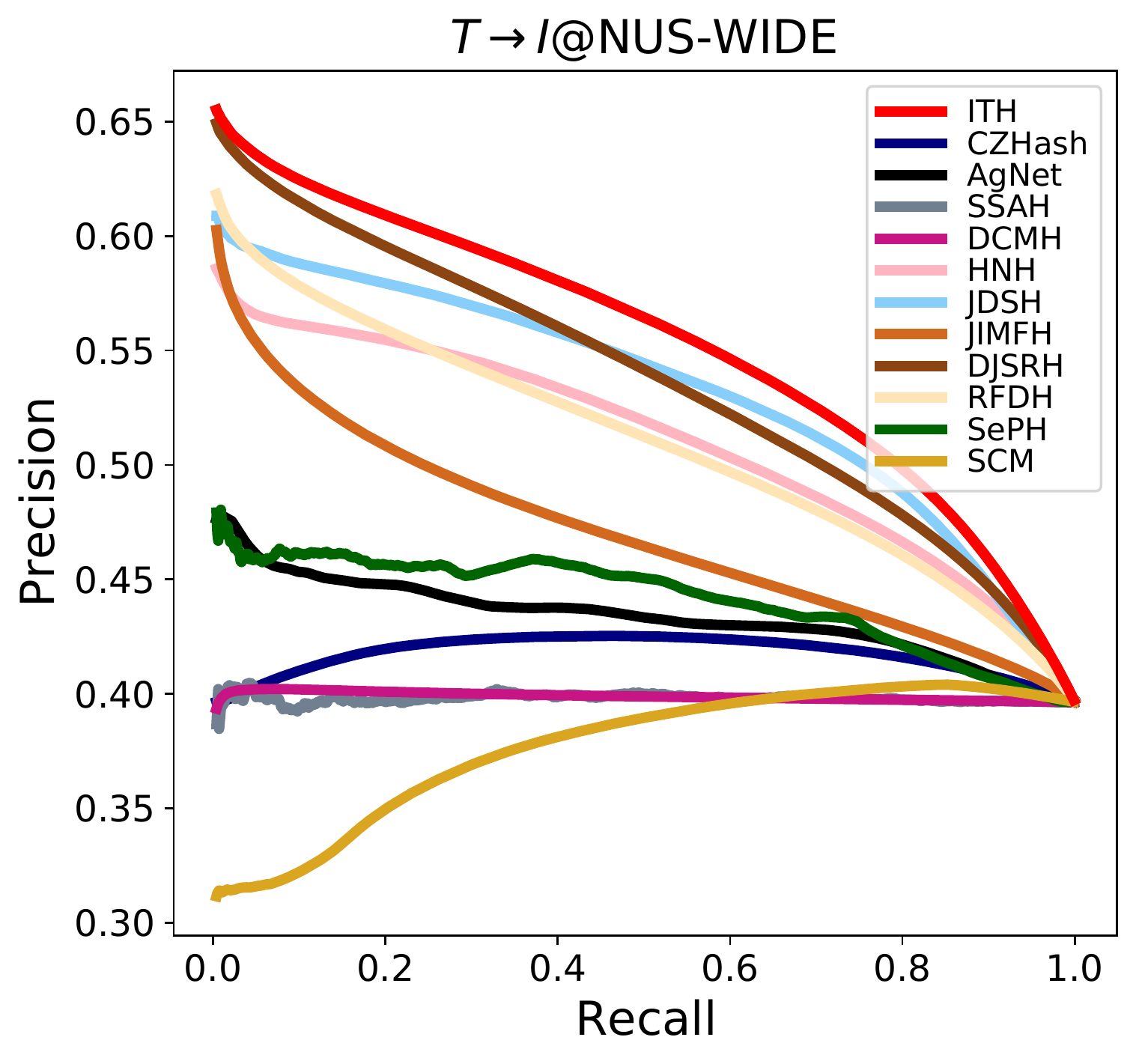}
\end{minipage}%
}%
\centering
\caption{Precision-Recall curves with 128 bits on three benchmark datasets.} \label{fig_tnpr}
\end{figure}
Along with the retrieval accuracy, the query speed is also a critical factor. To achieve constant or sub-linear retrieval speed, hamming-based zero-shot cross-modal retrieval is conducted in terms of the hash lookup protocol. Accordingly, by changing the hamming distance radius from 0 to 128 bits, the PR curves of the proposed ITH and binary-value baselines are plotted in Fig.~\ref{fig_tnpr}. For PR curves, the higher location denotes better accuracy. Therefore, ITH obtains competitive results when compared with other methods in Fig.~\ref{fig_tnpr}. Along with the intrinsic semantics of multi-modal data, the consideration of the total correlation among each bit also boosts the accuracy. Since the bit-wise correlation is reduced, the resolution ratio of hamming distance is correspondingly increased, which promotes the separation of data points with different semantics. Consequently, ITH achieves compelling performance in terms of hash lookup protocol.

\subsection{Ablation Study}\label{ablation_study}
To validate the contribution of elements in ITH, ablation study is implemented on the Wikipedia, Pascal Sentences and NUS-WIDE datasets with 128 bits. Specifically, (1)~the efficiency of the multi-modal intrinsic semantics and (2)~the contribution of the modules in ITH are investigated.

\begin{table}[!h]
\caption{MAP of single modality retrieval on three benchmark datasets.}
\centering
\begin{tabular}{|c |c |c|c |}
\cline{1-4}
Source & Wikipedia &  Pascal Sentences & NUS-WIDE\\ 
\cline{1-4}
{\makecell{$F^v$ from \\pretrained VGG-19}} &0.3578	&0.5171	&0.5670\\
\cline{1-4}
{\makecell{$F^t$ from \\pretrained Doc2vec}} &0.5060	&0.4931	&0.6027\\
\cline{1-4}
{\makecell{$G^v$ via \\ ration in AIA}}& 0.3664	&0.5440	&0.6273\\
\cline{1-4}
{\makecell{$G^t$ via \\ ration in AIA}} & 0.5600	&0.4534	&0.6061\\
\cline{1-4}
{\makecell{$G^z$ via\\ fusion in AIA}} &\textbf{0.4108}	&\textbf{0.5673}	&\textbf{0.6438}\\
\cline{1-4}
\end{tabular}
\label{tab6}
\end{table}

\subsubsection{The multi-modal intrinsic semantics} The motivation of ITH is to reduce the modality gap by exploring the semantics of multi-modal data with no extra information. Therefore, the Adaptive Information Aggregation~(AIA) in ITH aggregates the semantics of pre-trained features~(i.e., $\boldsymbol{F}^v$ and $\boldsymbol{F}^t$). Specifically, the ration sub-module transforms $\boldsymbol{F}^v$ and $\boldsymbol{F}^t$ into the refined representations for target datasets~(i.e., $\boldsymbol{G}^v$ and $\boldsymbol{G}^t$) for quantification, and the fusion sub-module further aggregates the multi-modal semantics as the fused vector~(i.e., $\boldsymbol{G}^z$). 

To evaluate the semantics of above-mentioned features, the similarity of multi-modal instances are measured with the cosine similarity in different spaces, and the corresponding MAPs are reported in Table~\ref{tab6}. As can be seen, the pretrained features $\boldsymbol{F}^v$ and $\boldsymbol{F}^t$ can bring good retrieval performance in single modality, which indicates that the pretrained features contain rich semantics. Meanwhile, compared with $\boldsymbol{F}^v$ and $\boldsymbol{F}^t$, the refined representations $\boldsymbol{G}^v$ and $\boldsymbol{G}^t$ are generally more suitable for target datasets. It verifies that the adjustment of features is needed to fit down-stream tasks. Finally, by adaptively aggregating the multi-modal semantics, the fused vector $\boldsymbol{G}^z$ usually achieves the best performance. The only exception is that the MAP of $\boldsymbol{G}^z$ is lower than $\boldsymbol{G}^t$ on the Wikipedia dataset. One possible reason is that the gap between image and text is much larger in describing the abstract concepts. Even so, the retrieval performance of fused information is still better than images.

\begin{table*}[!h]
\caption{MAP of the proposed ITH with different similarity guidance on three benchmark datasets.}
\centering
\begin{tabular}{|c |c  c c |c c  c |c c  c|}
\cline{1-10}
\multirow{2}*{Method} & \multicolumn{3}{c|}{Wikipedia} &  \multicolumn{3}{c|}{Pascal Sentences} &  \multicolumn{3}{c|}{NUS-WIDE}\\ 
\cline{2-10}& $I\rightarrow T$ & $T\rightarrow I$ & Avg & $I\rightarrow T$ & $T\rightarrow I$ & Avg & $I\rightarrow T$ & $T\rightarrow I$ & Avg\\
\cline{1-10}
Original Image: $F^v$ &0.3389	&0.3142	&0.3265	&0.4665	&0.4137	&0.4401	&0.5312	&0.5382	&0.5347\\
Original Text: $F^t$ &0.3238	&0.3172	&0.3205	&0.4322	&0.4141	&0.4231	&0.4443	&0.4697	&0.4570\\
Original Class: $F^y$ &0.2983	&0.2608	&0.2795	&0.3595	&0.3183	&0.3389	&0.3956	&0.3956	&0.3956\\
\cline{1-10}
Refined Image: $G^v$ &0.3624	&0.3377	&0.3501	&0.4700	&0.4378	&0.4539	&0.5633	&0.5584	&0.5608\\
Refined Text: $G^t$  &0.3305	&0.3184	&0.3244	&0.4618	&0.4404	&0.4511	&0.5592	&0.5656	&0.5624\\
\cline{1-10}
Fused representation: $G^z$ & \textbf{0.3656}	& \textbf{0.3403}	& \textbf{0.3529}	& \textbf{0.4755}	& \textbf{0.4630}	& \textbf{0.4692}	& \textbf{0.5761}	& \textbf{0.5696}	& \textbf{0.5728}\\ 
\cline{1-10}
\end{tabular}
\label{tab4}
\end{table*}

The semantics of features in different stages are further evaluated in cross-modal retrieval. Specifically, their cosine similarity is utilized as the semantic relations of common space, which guides the hash codes learning in SPE. In Table~\ref{tab4}, MAP of the proposed ITH with different similarity guidance are reported. From Table~\ref{tab4}, two phenomena can be observed. Firstly, with the same relations as guidance, compared with the performance of single modality in Table~\ref{tab6}, the accuracy of cross-modal retrieval reduces due to the modality gap. Secondly, the common space guided by the fusion relations $\boldsymbol{G}^z$ can achieve better performance than the relations in the single modality. It demonstrates that the biased relations are not feasible for processing multi-modal data. Consequently, the efficiency of the multi-modal intrinsic semantics is verified.

\begin{table*}[!h]
\caption{MAP of ablation experiments on three benchmark datasets.}
\centering
\begin{tabular}{|c|c |c  c c |c c  c |c c  c|}
\cline{1-11}
\multirow{2}*{Stage} & \multirow{2}*{Method} & \multicolumn{3}{c|}{Wikipedia} &  \multicolumn{3}{c|}{Pascal Sentences} &  \multicolumn{3}{c|}{NUS-WIDE}\\ 
\cline{3-11}& & $I\rightarrow T$ & $T\rightarrow I$ & Avg & $I\rightarrow T$ & $T\rightarrow I$ & Avg & $I\rightarrow T$ & $T\rightarrow I$ & Avg\\
\cline{1-11}
\multirow{2}*{1 AIA Ration} & ITH-w/o 1-regularity &0.3534	&0.3337	&0.3435	&0.4602	&0.4374	&0.4488	&0.5628	&0.5670	&0.5649\\
&ITH-w/o 1-succession &0.3319	&0.3152	&0.3235	&0.3935	&0.3732	&0.3833	&0.4495	&0.4962	&0.4728\\
\cline{1-11}
\multirow{3}*{2 AIA Fusion} & ITH-w/o 2-regularity &0.3511	&0.3348	&0.3429	&0.4461	&0.4586	&0.4523	&0.5723	&0.5672	&0.5697\\
&ITH-w/o 2-succession &0.3110	&0.2829	&0.2969	&0.3679	&0.3755	&0.3717	&0.4453	&0.4455	&0.4454\\
&ITH-w/o 2-auto mix &0.3427	&0.3228	&0.3327	&0.4750	&0.4625	&0.4687	&0.5373	&0.5520	&0.5446\\
\cline{1-11}
\multirow{3}*{3 SPE Alignment} & ITH-w/o Intra &0.3423	&0.3241	&0.3332	&0.4612	&0.4341	&0.4476	&0.5564	&0.5640	&0.5602\\
&ITH-w/o Inter &0.2180	&0.2182	&0.2181	&0.1885	&0.1772	&0.1828	&0.3957	&0.3946	&0.3951\\
&ITH-w/o TC &0.3594	&0.3387	&0.3490	&0.4468	&0.4336	&0.4402	&0.5458	&0.5564	&0.5511\\
\cline{1-11}
\multicolumn{2}{|c|}{\textbf{ITH-FULL}}  & \textbf{0.3656}	& \textbf{0.3403}	& \textbf{0.3529}	& \textbf{0.4755}	& \textbf{0.4630}	& \textbf{0.4692}	& \textbf{0.5761}	& \textbf{0.5696}	& \textbf{0.5728}\\ 
\cline{1-11}
\end{tabular}
\label{tab5}
\end{table*}

\subsubsection{The modules in ITH} To confirm the contribution of modules in the proposed method, several variants of ITH are built and evaluated. Table~\ref{tab5} reports the results of ablation experiments on three benchmarks datasets with 128 bits. 

For the ration sub-module in AIA, based on the S-PRI, it builds refined representations to quantify the semantics of different modalities. ITH-w/o 1-regularity denotes the variant that the optimization of cross-entropy loss for $\textit{ImgFeaNet}$ and $\textit{TxtFeaNet}$ is interrupted. Meanwhile, ITH-w/o 1-succession is the ITH that eliminates the descriptive power objective in the ration module. The former controls the uncertainty reduction for target datasets, and the latter constrains the succession of pre-trained semantics in the refined representations. As can be seen, the elimination of above two factors leads to decrease, which demonstrates that both regularity and succession in the S-PRI are necessary. Furthermore, compared with the ITH-w/o 1-regularity, the decline of ITH-w/o 1-succession is more distinct. This phenomenon indicates that the semantics of pre-trained features is more generalized than the semantics brought by class labels. 

For the fusion sub-module in AIA, based on the S-PRI, it aggregates the quantified multi-modal semantics as the fused vectors. To verify the components in the fusion module, three variants are investigated:(1) Eliminate the class information and constrain the fused vectors to only grasp the quantified multi-modal pre-trained semantics~(i.e., ITH-w/o 2-regularity). (2) Remove the pre-trained semantics and utilize the fused vectors to only reveal the class information~(i.e., ITH-w/o 2-succession). (3) Remove the adaptive quantification for multi-modal distance relations~(i.e. Eq.~(\ref{eq10})), and set the equal weight for different modalities~(i.e., ITH-w/o 2-auto mix). As shown in Table~\ref{tab5}, the removal of components in the fusion sub-module inevitably causes the decrease of ITH. Meanwhile, for the regularity and succession, their absence in the fusion sub-module is more destructive than that in the ration sub-module. It is because that the guidance for subsequent hash code learning is provided by the fusion sub-module. Moreover, the impact of semantics quantification for different datasets is different. It demonstrates that for large-scale training set, the semantic guidance should be more reliable. 

\begin{figure}[!h]
\centering
\subfigure[Wikipedia]{
\begin{minipage}[t]{0.49\linewidth}
\centering
\includegraphics[width=1.7in]{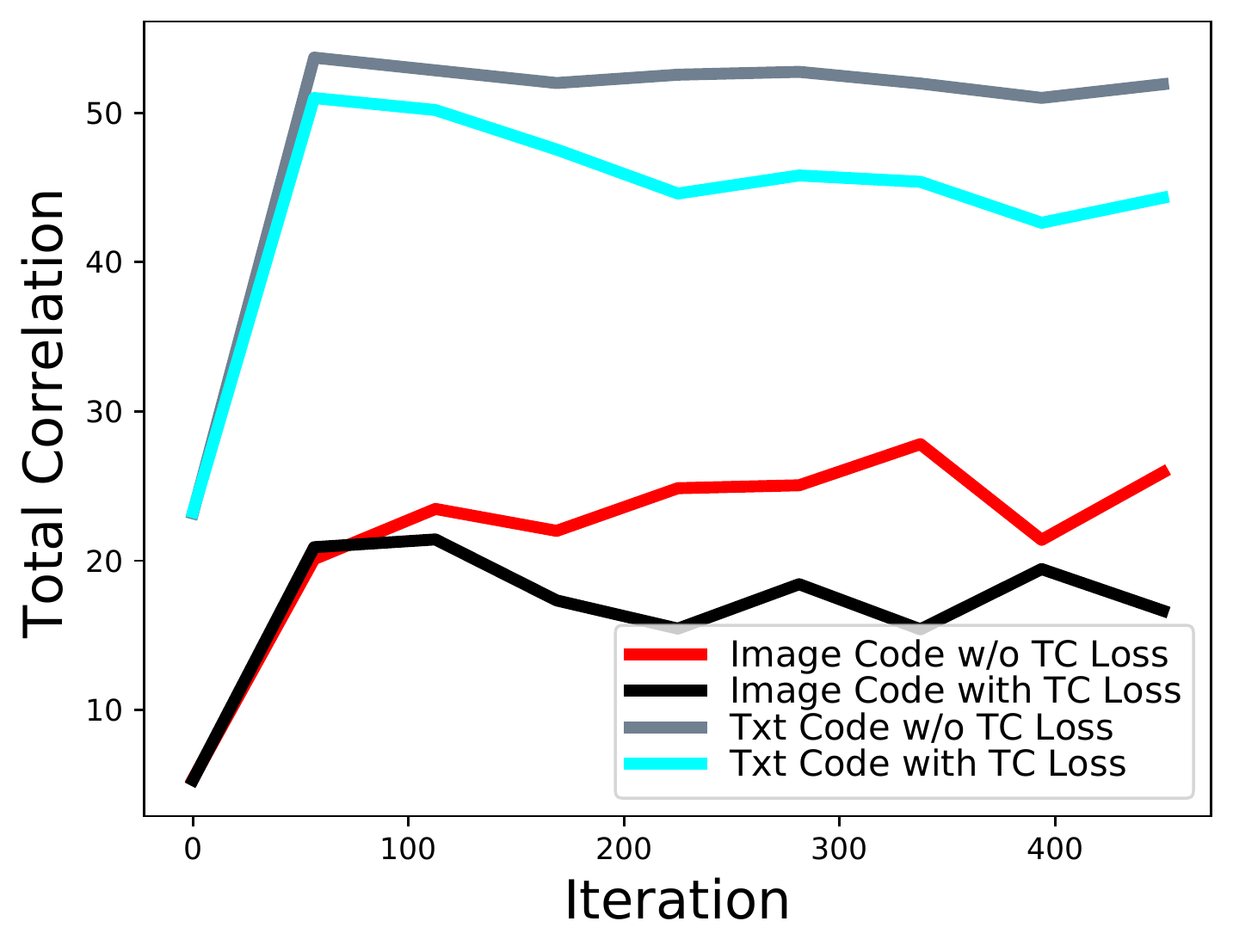}
\end{minipage}%
}%
\subfigure[NUS-WIDE]{
\begin{minipage}[t]{0.49\linewidth}
\centering
\includegraphics[width=1.7in]{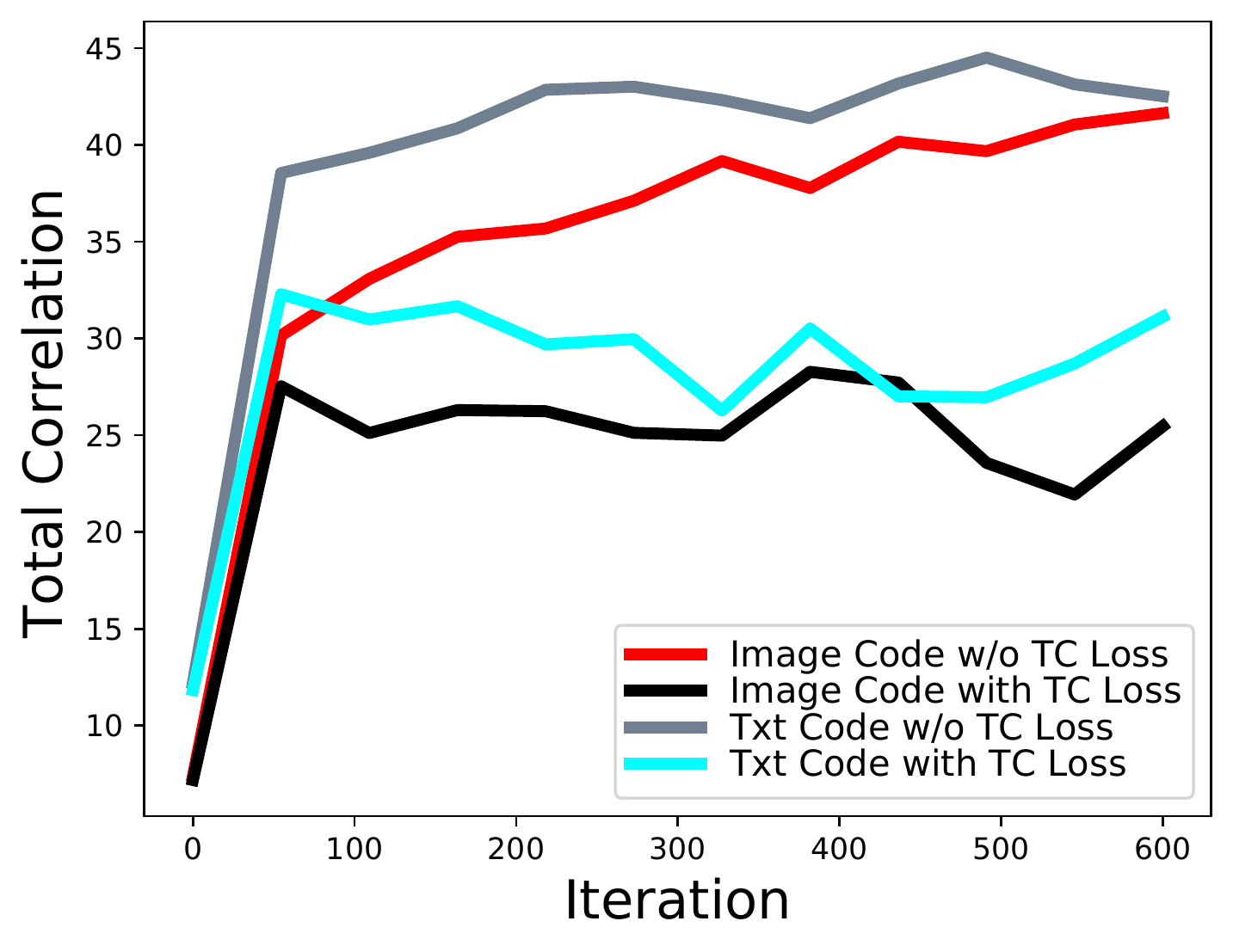}
\end{minipage}%
}%
\centering
\caption{Total Correlation~(TC) on Wikipedia and NUS-WIDE datasets with 128 bits.}
\label{fig_tc}
\end{figure}

For the SPE module, guided by the intrinsic semantics, it encodes data points as hash codes. As introduced in Section~\ref{sec_SPE}, hash codes should reveal the intra-modal correlation and the inter-modal similarity. Meanwhile, the bit-wise correlation of hash codes is also decoupled via the TC loss. Correspondingly, three variants are implemented: ITH-w/o 3 Intra, ITH-w/o 3 Intra and ITH-w/o 3 TC. As shown in Table~\ref{tab5}, the inter-similarity similarity is more important than the intra-modal correlation. Therefore, to boost retrieval accuracy, the similarity guidance should give more attention. Meanwhile, the elimination of TC loss also leads to a drop, which indicates that the representation ability of hash codes is improved by the TC loss. To further evaluate the TC loss, the total correlation of hash codes is visualized in Fig.~\ref{fig_tc}. As can be seen, the optimization of TC loss indeed decreases the bit-wise correlation.



 








\subsection{Parameter Analysis}\label{paramerter_test}
Finally, to explore the influence of hyper-parameters~(i.e., $\alpha$, $\beta$, $\gamma$ and $\eta$, parameter experiments are conducted on the Pascal Sentences and Wikipedia datasets with 16 bits as code length. Specifically, the analysis for one parameter is implemented by changing its value and keeping other parameters fixed as experimental settings~(i.e., $\alpha=100,\beta=100,\gamma=1, \eta=0.01$). Since the proposed ITH mainly consists of the ration sub-module in AIA, the fusion sub-module in AIA, and SPE module, the analysis of parameters is carried out according to the affiliation. 

\begin{figure}[!h]
\centering
\subfigure[$I\rightarrow T$]{
\begin{minipage}[t]{0.49\linewidth}
\centering
\includegraphics[width=1.7in]{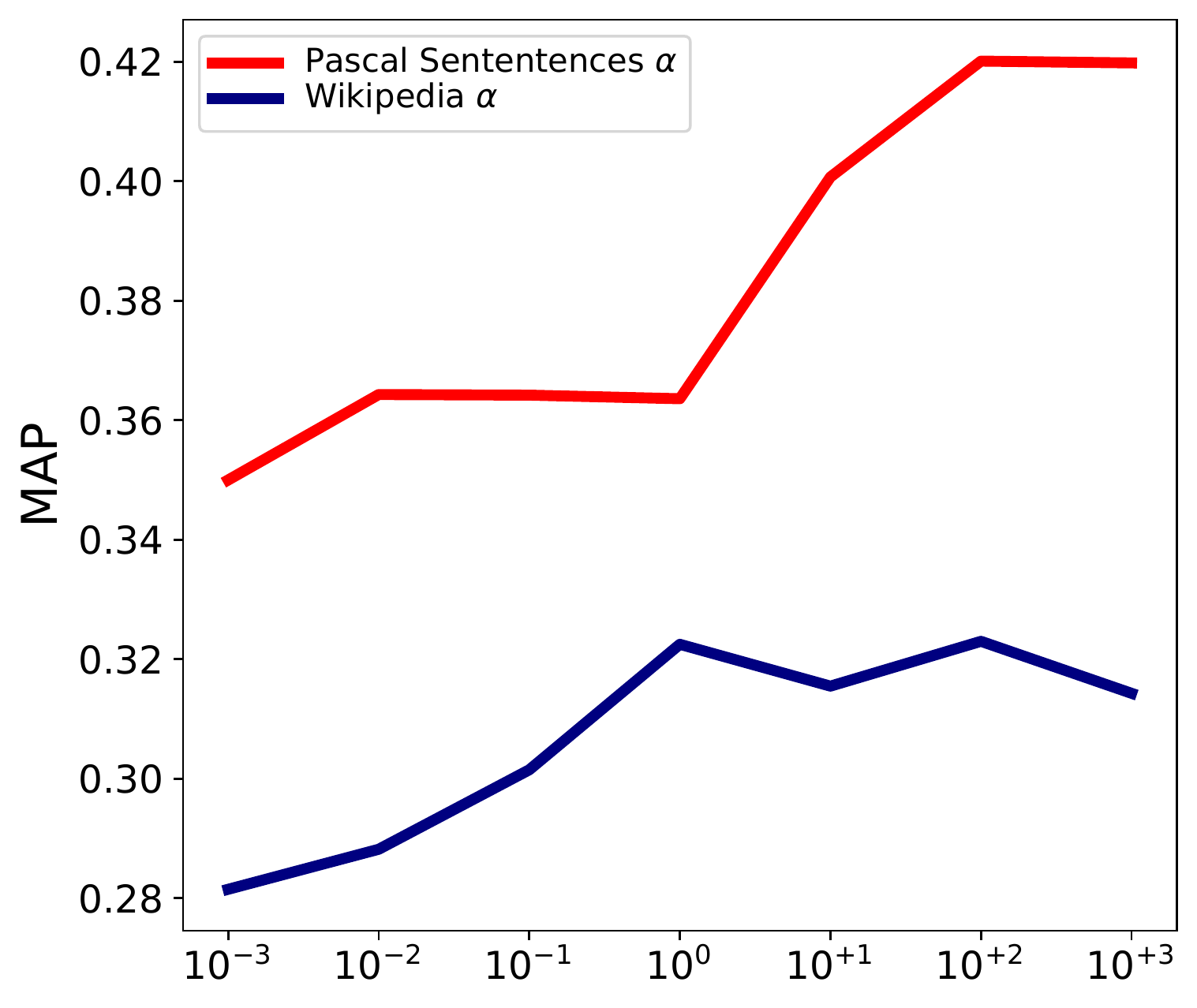}
\end{minipage}%
}%
\subfigure[$T\rightarrow I$]{
\begin{minipage}[t]{0.49\linewidth}
\centering
\includegraphics[width=1.7in]{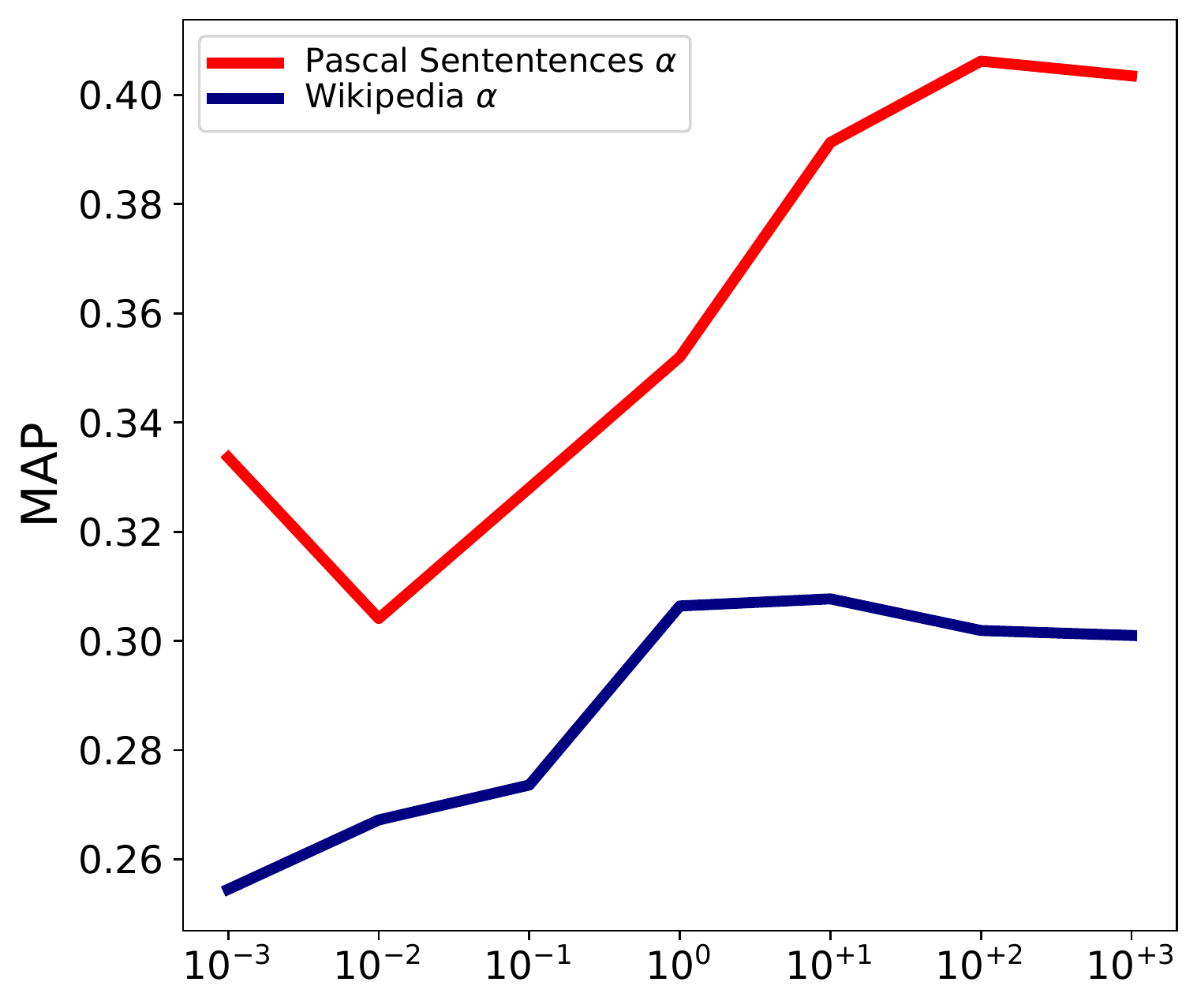}
\end{minipage}%
}%
\centering
\caption{MAP of the proposed ITH with different $\alpha$ values on two benchmark datasets.}
\label{fig_alpha}
\end{figure}

The ration sub-module in AIA quantifies the semantics of different modalities based on the corresponding refined representations. To build refined representations for different modalities, the hyper-parameter $\alpha$ controls the inheritance of the pre-trained semantics. Fig.~\ref{fig_alpha} illustrates the MAP of ITH with different $\alpha$. From Fig.~\ref{fig_alpha}, larger $\alpha$~(i.e, $\alpha \in \left[ 10^1,10^2 \right] $) will boost the performance of ITH, which highlights the importance of the pre-trained semantics of different modalities.

\begin{figure}[!h]
\centering
\subfigure[$I\rightarrow T$]{
\begin{minipage}[t]{0.49\linewidth}
\centering
\includegraphics[width=1.7in]{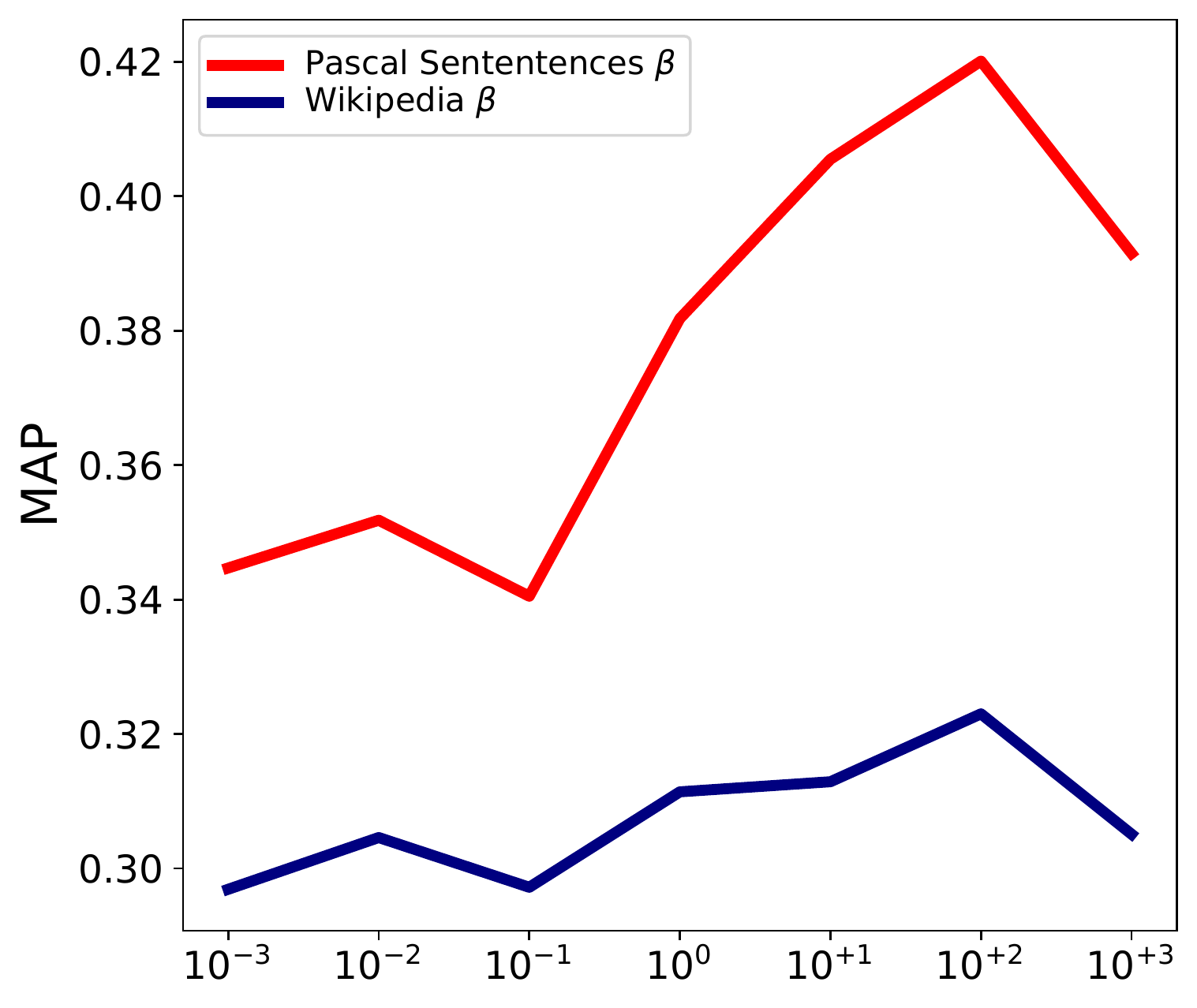}
\end{minipage}%
}%
\subfigure[$T\rightarrow I$]{
\begin{minipage}[t]{0.49\linewidth}
\centering
\includegraphics[width=1.7in]{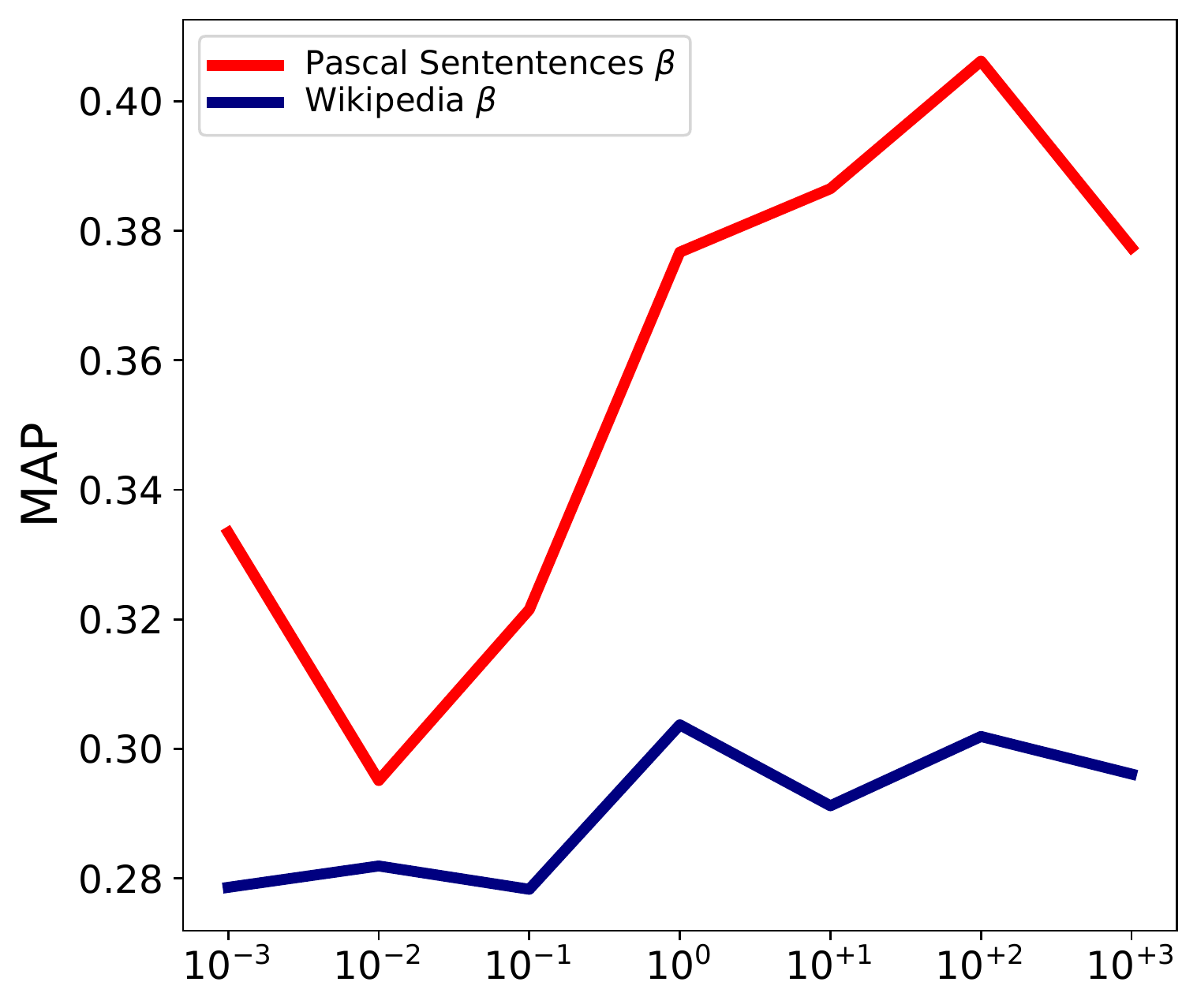}
\end{minipage}%
}%
\centering
\caption{MAP of the proposed ITH with different $\beta$ values on two benchmark datasets.}
\label{fig_beta}
\end{figure}

\begin{figure}[!htp]
\centering
\includegraphics[scale=0.3]{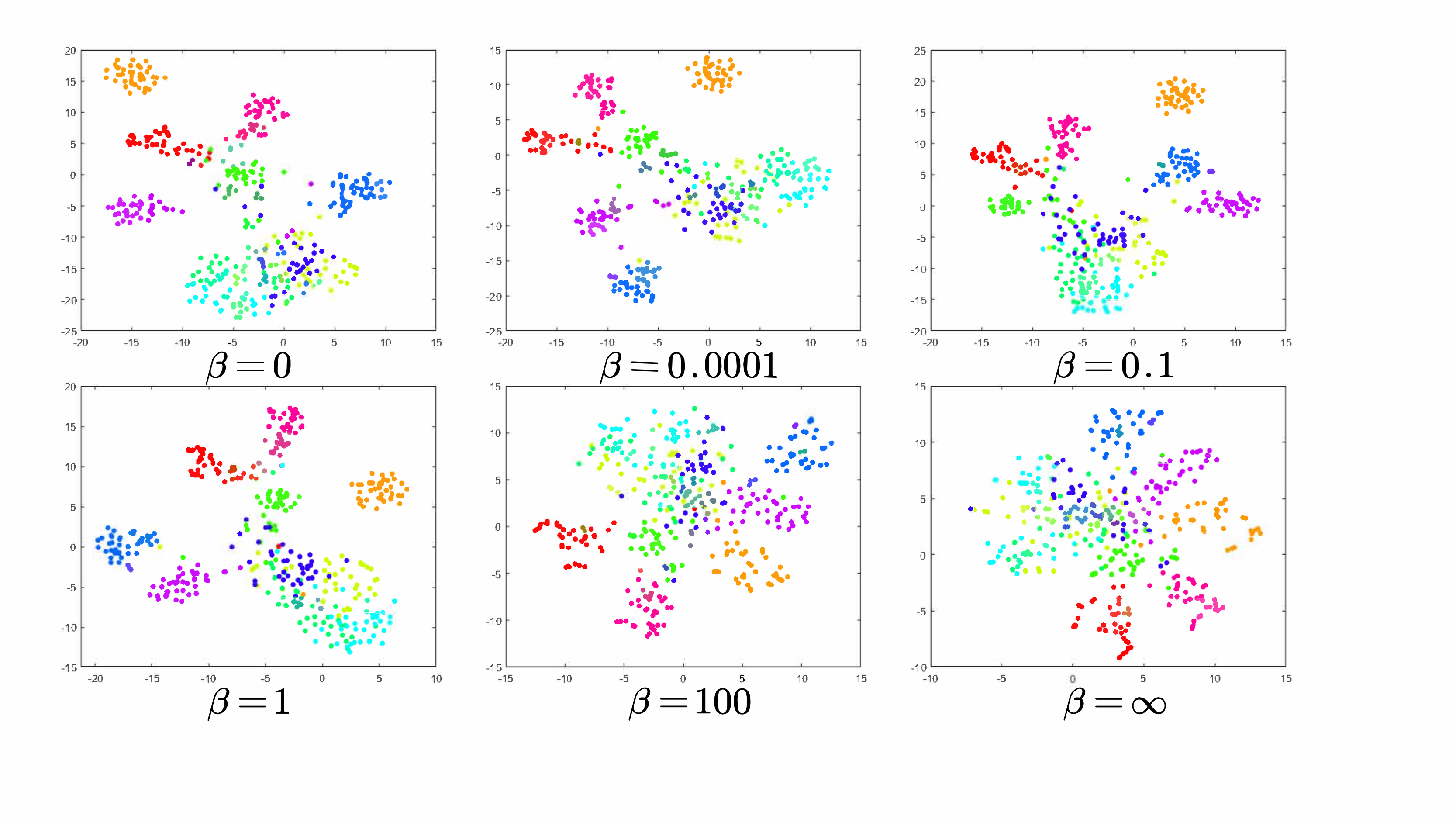}
\caption{T-SNE visualization of $\boldsymbol{G}^z$ for the training set of Pascal Sentence dataset with different $\beta$ in the S-PRI.}\label{fig_priadd}
\end{figure}

The fusion sub-module in AIA reveals the intrinsic semantics of multi-modal data with the fused representation $\boldsymbol{G}^z$. Based on the S-PRI, the hyper-parameter $\beta$ balances the inheritance of the pre-trained semantics and the semantic uncertainty reduction. To analyze the influence of $\beta$, the distributions of $\boldsymbol{G}^z$ with different $\beta$ are visualized in Fig.~\ref{fig_priadd}. Meanwhile, the MAP of ITH with different $\beta$ is also plotted in Fig.~\ref{fig_beta}. With the $\beta$ increases, the inheritance of semantics is gradually enhanced, and the semantic uncertainty reduction decays. This observation matches well with the effects of the trade-off parameter as demonstrated in Fig.~\ref{fig_pri_synthetic}.
According to  Fig.~\ref{fig_beta},
$\beta$ should be set in a reasonable range (i.e., $\beta \in \left[ 10^0,10^2 \right]$ ). A relative large (e.g., $\beta \leq 10^3$) or small (e.g., $\beta \leq 10^{-1}$) value of $\beta$ may hurt the performance.

\begin{figure}[!h]
\centering
\subfigure[$I\rightarrow T$]{
\begin{minipage}[t]{0.49\linewidth}
\centering
\includegraphics[width=1.7in]{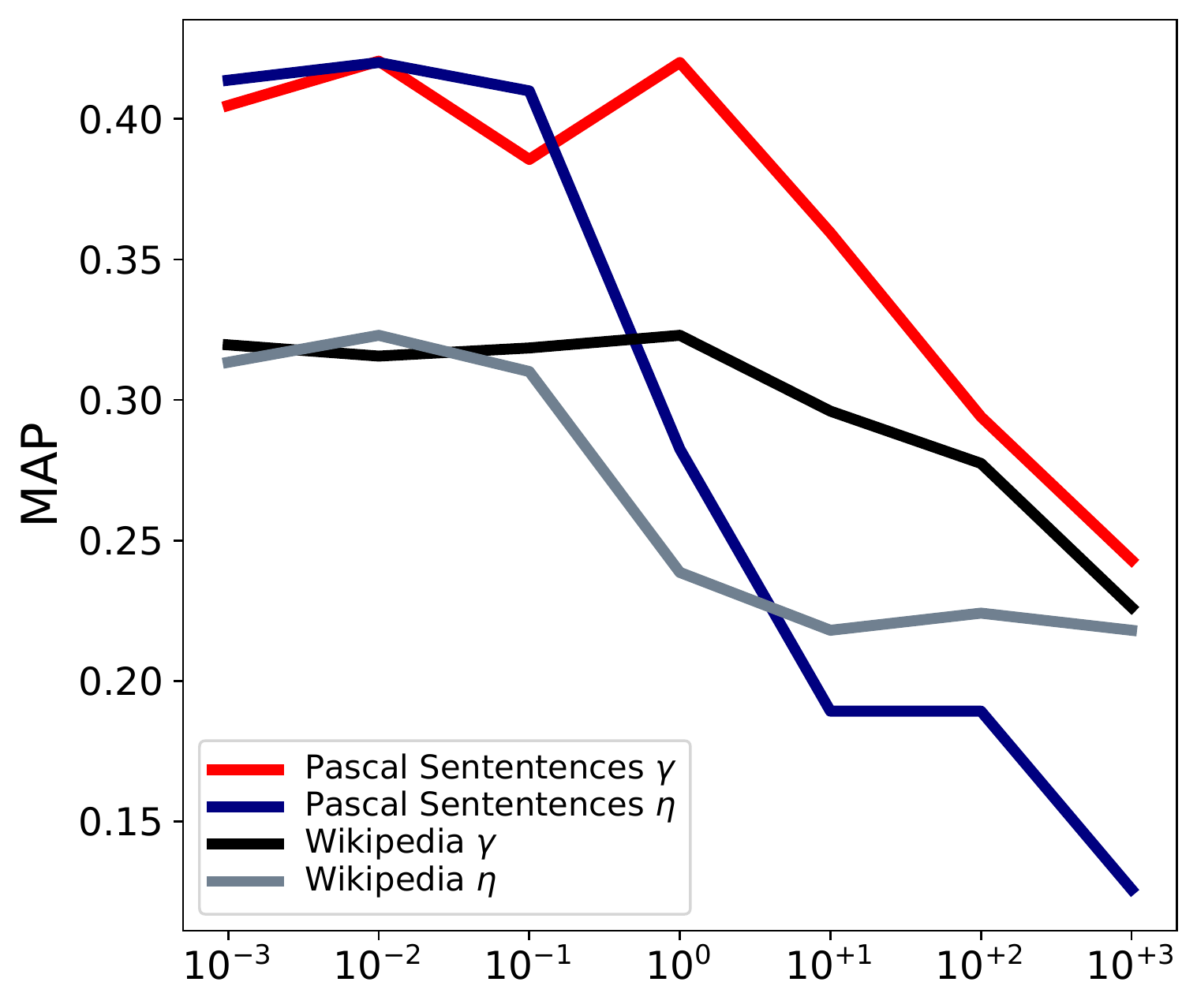}
\end{minipage}%
}%
\subfigure[$T\rightarrow I$]{
\begin{minipage}[t]{0.49\linewidth}
\centering
\includegraphics[width=1.7in]{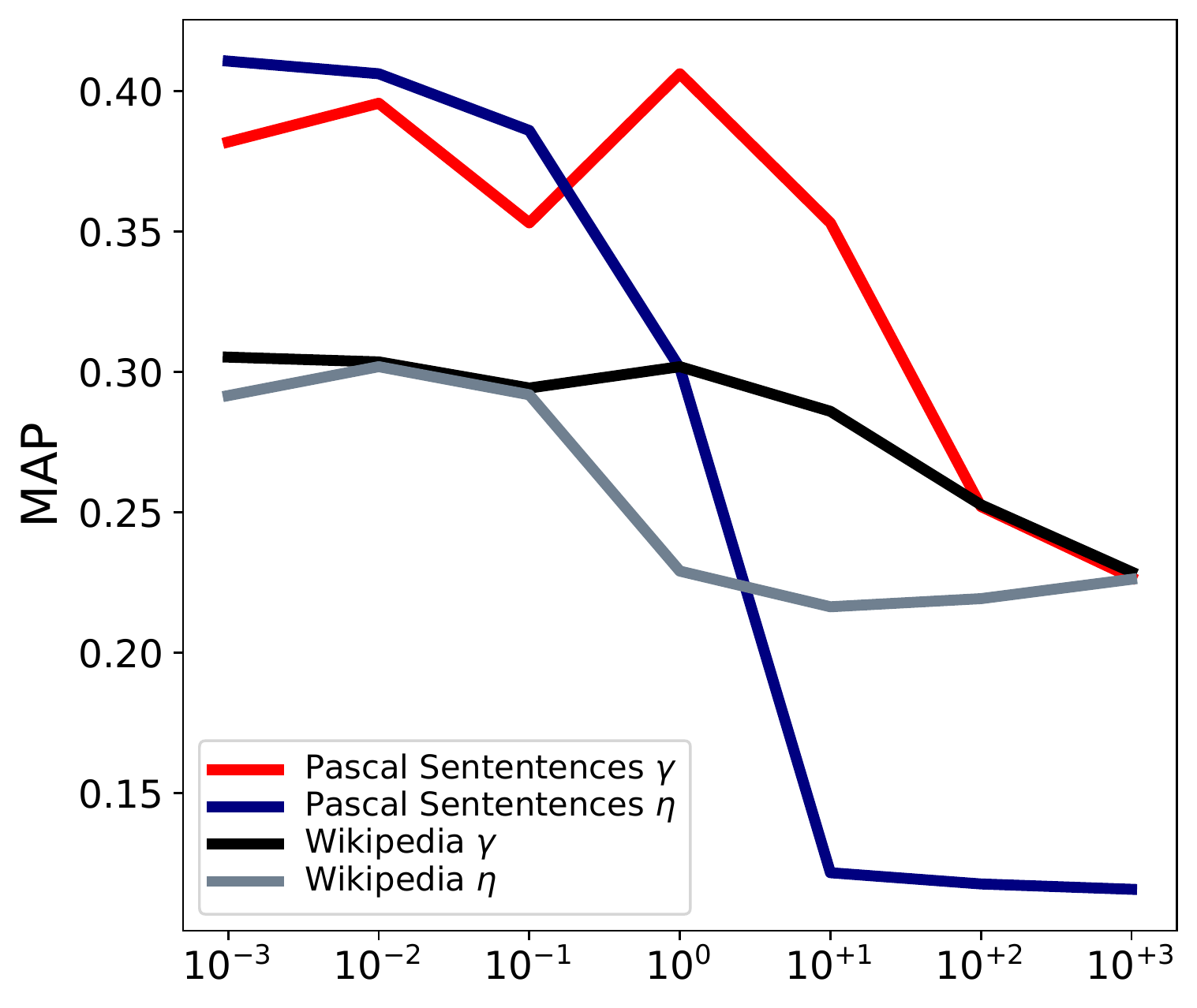}
\end{minipage}%
}%
\centering
\caption{MAP of the proposed ITH with different $\gamma$ and $\eta$ values on two benchmark datasets.}
\label{fig_gamma_eta}
\end{figure}

By preserving the intrinsic semantics of multi-modal data, the SPE module optimizes the $\textit{ImgHashNet}$ and $\textit{TxtHashNet}$ to encode data points as hash codes. Fig.~\ref{fig_gamma_eta} shows the MAP of ITH with different weights in the objective function of SPE~(i.e., Eq.~(\ref{eq20})). During the optimization of SPE, $\gamma$ influences the mixture of the inter-modal semantics preservation and the intra-modal correlation. As shown in Fig.~\ref{fig_gamma_eta}, ITH can achieve the satisfying accuracy when $\gamma$ is smaller than 1. It indicates that the inter-modal semantics preservation is more important than the intra-modal correlation. Meanwhile, $\eta$ adjusts the
Total Correlation loss, which aims to improve the representation ability of hash codes. It is observed that $\eta$ should be assigned with a smaller value~(i.e., $\eta \in \left[ 10^{-3},10^{-1} \right] $). It is because that the preservation of the intrinsic semantics has priority over the extra requirement for hash codes. Based on above results, ITH is robust to the changes of the hyper-parameters in a reasonable range.

\section{Conclusion}\label{sec5}
This paper presents a novel hashing method named Information-Theoretic Hashing~(ITH) for zero-shot cross-modal retrieval. Within the proposed method, the adaptive information aggregation module follows the Principle of Relevant Information~(PRI) to quantify and aggregate the semantics of different modalities, whereby the intrinsic semantics of multi-modal data is revealed. Instead of utilizing auxiliary information to guarantee generalization, the semantics preserving hashing module encodes the multi-modal intrinsic semantics to build the common hamming space. In addition, the total correlation loss is designed to address the bit-wise correlation of hash codes. Sufficient experiments on three public datasets suggest that the common space built by ITH is effective for reducing the modality gap, and demonstrate the improved retrieval accuracy in comparison with state-of-the-arts.


\ifCLASSOPTIONcaptionsoff
  \newpage
\fi

\bibliographystyle{IEEEtran}
\bibliography{mylib}








\end{document}